\documentclass[10pt, conference, compsocconf]{IEEEtran}
\usepackage{cite}
\usepackage{amsmath,amssymb,amsfonts}
\usepackage{algorithmic}
\usepackage{graphicx}
\usepackage{textcomp}
\usepackage{xcolor}
\def\BibTeX{{\rm B\kern-.05em{\sc i\kern-.025em b}\kern-.08em
    T\kern-.1667em\lower.7ex\hbox{E}\kern-.125emX}}

\usepackage{epsfig}
\usepackage{graphicx}
\usepackage{amsmath}
\usepackage{amssymb}
\usepackage{bm}

\usepackage{soul}
\usepackage[font=small,labelfont=bf]{caption}
\usepackage{booktabs}
\usepackage{mdframed}

\usepackage{subcaption}
\usepackage{pdflscape}
\usepackage{multirow}
\usepackage{color}
\usepackage[normalem]{ulem}

\usepackage[pagebackref=true,breaklinks=true,letterpaper=true,colorlinks,bookmarks=false]{hyperref}

\ifCLASSINFOpdf
\else
\fi
\begin{document}
%
\title{Ordinal-Content VAE:\\ 
Isolating Ordinal-Valued Content Factors in Deep Latent Variable Models}


\author{\IEEEauthorblockN{Minyoung Kim$^1$}
\IEEEauthorblockA{$^1$Samsung AI Center\\
Cambridge, UK\\
mikim21@gmail.com}
\and
\IEEEauthorblockN{Vladimir Pavlovic$^{1,2}$}
\IEEEauthorblockA{
$^2$Rutgers University\\
Piscataway, NJ, USA\\
vladimir@cs.rutgers.edu}
}


%


\maketitle

\begin{abstract}
In deep representational learning, it is often desired to isolate a particular factor (termed {\em content}) from other factors (referred to as {\em style}). What constitutes the content is typically specified by users through explicit labels in the data, while all unlabeled/unknown factors are regarded as style. Recently, it has been shown that such content-labeled data can be effectively exploited by modifying the deep latent factor models (e.g., VAE) such that the style and content are well separated in the latent representations. However, the approach assumes that the content factor is categorical-valued (e.g., subject ID in face image data, or digit class in the MNIST dataset). In certain situations, the content is ordinal-valued, that is, the values the content factor takes are {\em ordered} rather than categorical, making content-labeled VAEs, including the latent space they infer, suboptimal. In this paper, we propose a novel extension of VAE that imposes a partially ordered set (poset) structure in the content latent space, while simultaneously making it aligned with the ordinal content values. To this end, instead of the iid Gaussian latent prior adopted in prior approaches, we introduce a conditional Gaussian spacing prior model. This model admits a tractable joint Gaussian prior, but also effectively places negligible density values on the content latent configurations that violate the poset constraint. To evaluate this model, we consider two specific ordinal structured problems: estimating a subject's age in a face image and elucidating the calorie amount in a food meal image.  We demonstrate significant improvements in content-style separation over previous non-ordinal approaches.
\end{abstract}


%
\IEEEpeerreviewmaketitle

\section{Introduction}\label{sec:intro}

Identifying underlying sources of variability that explain 
the generative process of high dimensional data 
is one of the key problems in recent deep representation learning. One of the main objectives is to learn proper latent vector representation for these factors, leading to latent representation that is succinct, faithfully reconstructs the original data, and disentangles different factors from each other~\cite{bengio_disent}. 

The unsupervised representation learning ambitiously aims to learn the factors solely from the observed data 
without any supervision~\cite{infogan16,aae16,beta_vae17,nica17,dip_vae18,factor_vae18,tcvae,disent19}. However, most of these approaches face the inherent impossibility~\cite{impossibility} of identifying the unsupervised factors. To alleviate this challenge, several weakly or semi-supervised learning frameworks have been proposed~\cite{sup_reed,sup_yang,sup_kulkarni,sup_whitney,lecun_disent,DBLP:journals/corr/abs-1901-08534}. Among those, \cite{mlvae} proposes a content-style separation framework, where the goal is to isolate a particular factor (termed {\em content}) from remaining factors (referred to as {\em style}). What constitutes the content is typically specified by users through explicit labels in the data, while all unlabeled/unknown factors are regarded as style.

Under the setting of content labels for the data, more specifically in the form of paired data $({\bf x},c)$ where ${\bf x}$ is a data instance (e.g., image) and $c\in\{1,\dots,K\}$ is its content label taking $K$ different levels, it would be natural to define and learn $K$ latent content vectors $\{{\bf v}_i\}_{i=1}^K$, one for each content value, namely ${\bf v}_1$ for $c=1$, ${\bf v}_2$ for $c=2$, and so on. The group-level VAE model proposed in~\cite{mlvae} (denoted here as ML-VAE) is an extension of VAE~\cite{vae14} with the modeling assumption that each instance ${\bf x}$ with content value $c=i$ is generated from both the content latent vector ${\bf v}_i$ and the instance-specific style latent vector ${\bf s}$. This implies that all the instances ${\bf x}$ with the same content value $c=i$,  
share the content latent vector ${\bf v}_i$ whereas the latent vector ${\bf s}$ captures the remaining factor variations specific to each instance. The proposed modified variational learning encourages style and content to be well separated in the latent representations, which is empirically demonstrated on the applications of isolating subject ID (content) in face images from the other factors (e.g., facial pose and expression), and the digit class (content) in handwritten digit images from e.g., writing style.


Unlike ML-VAE, we consider a different setup where the content factor is {\em ordinal-valued}, i.e., the values the content factor takes are {\em ordered} rather than categorical. For instance, a subject's age in a face image represents such an ordinal quantity; the calorie amount of a food meal, represented by its image, is naturally ordinal-valued. If the content factor $c$ has such an ordinal structure $c \in \{ 1 < 2 < \cdots < K \}$, it is important to embed this structure as we form the content space. Specifically, for two instances ${\bf x}$ and ${\bf x}'$ whose content values $c$ and $c'$ are close to each other, their respective content latent vectors ${\bf v}$ and ${\bf v}'$ should also be proximal, and vice versa. More formally, for any triplet $i<j<k$,
\begin{equation}
||{\bf v}_i - {\bf v}_k|| > \max\{ ||{\bf v}_i - {\bf v}_j||, ||{\bf v}_j - {\bf v}_k|| \}
\label{eq:content-distance-ineq}
\end{equation}
This condition will establish a good alignment between the content latent vectors and the ordinal content values, thus yielding a model closer to the true data generation process.

However, ML-VAE assumes the content factor is categorical, making it suboptimal in the ordinal-content setting. Specifically, they assume 
any content class (e.g., $c=1$) is equally different from all other classes (e.g., $c=2$ or $c=8$). In ordinal setting, $c=1$ is closer to $c=2$ than $c=8$. 
Enforcing this constraint in an effective and principled manner can be challenging: 
Directly incorporating the triplet inequalities as regularization loss into the  
objective function of  ML-VAE is 
heuristic, and it never guarantees that the ordinal constrains (\ref{eq:content-distance-ineq}) are satisfied in the embedded space.

In this paper we propose a new principled approach to ordinal content VAE learning by constructing an appropriate prior for the labeled content space. This is in contrast to the iid standard normal prior in the non-ordinal models like ML-VAE.  The prior is constructed by imposing a restrictive partially ordered set (poset) constraint, encoded in a novel conditional Gaussian spacing model. This prior assigns negligible density to the configurations of content vectors that violate the poset constraint. A key benefit of this model is that the joint prior becomes Gaussian, albeit with a full covariance, maintaining the closed form of the KL divergence term in the variational objective. Still the prior is fully factorized over the latent dimensions, leading to a computationally tractable model. Moreover, the number of parameters in the proposed prior 
is only $O(d \cdot K)$, where $d$ is the dimensionality of the content latent vector, 
reinforcing the model’s tractability.
We test our approach on both synthetic and real datasets, including the age of the subject as content in face images and the calorie amount as content in pizza images. Our 
model achieves significant improvement over the previous non-ordinal approaches in content-style separation, both quantitatively and qualitatively.

\section{Ordinal Content Level VAE 
}\label{sec:olvae}

\textbf{Setup and Goal}. Our proposed Ordinal Content Level VAE (OL-VAE) model, is an extension of VAE~\cite{vae14} and the ML-VAE for dealing with {\em ordinal-valued} content factors. We assume a semi-supervised setup with the content-annotated paired data $({\bf x},c)$. The content $c$ is discrete and 
ordinal valued: $1 < \cdots < K$. All other factors of variation are regarded as {\em style}, and we assume the style factors are not labeled. Our goal is to learn the latent representation, a pair of vectors $({\bf v},{\bf s})$ where ${\bf v}$ is responsible for content and ${\bf s}$ for style encoding, such that they are well separated and disentangled. That is, the change of ${\bf v}$ exclusively affects the content aspect and, conversely, the style latent ${\bf s}$ is independent of the content.

\begin{figure}
\vspace{-1.0em}
\begin{center}
\includegraphics[trim = 0mm 0mm 0mm 0mm, clip, scale=0.235
]{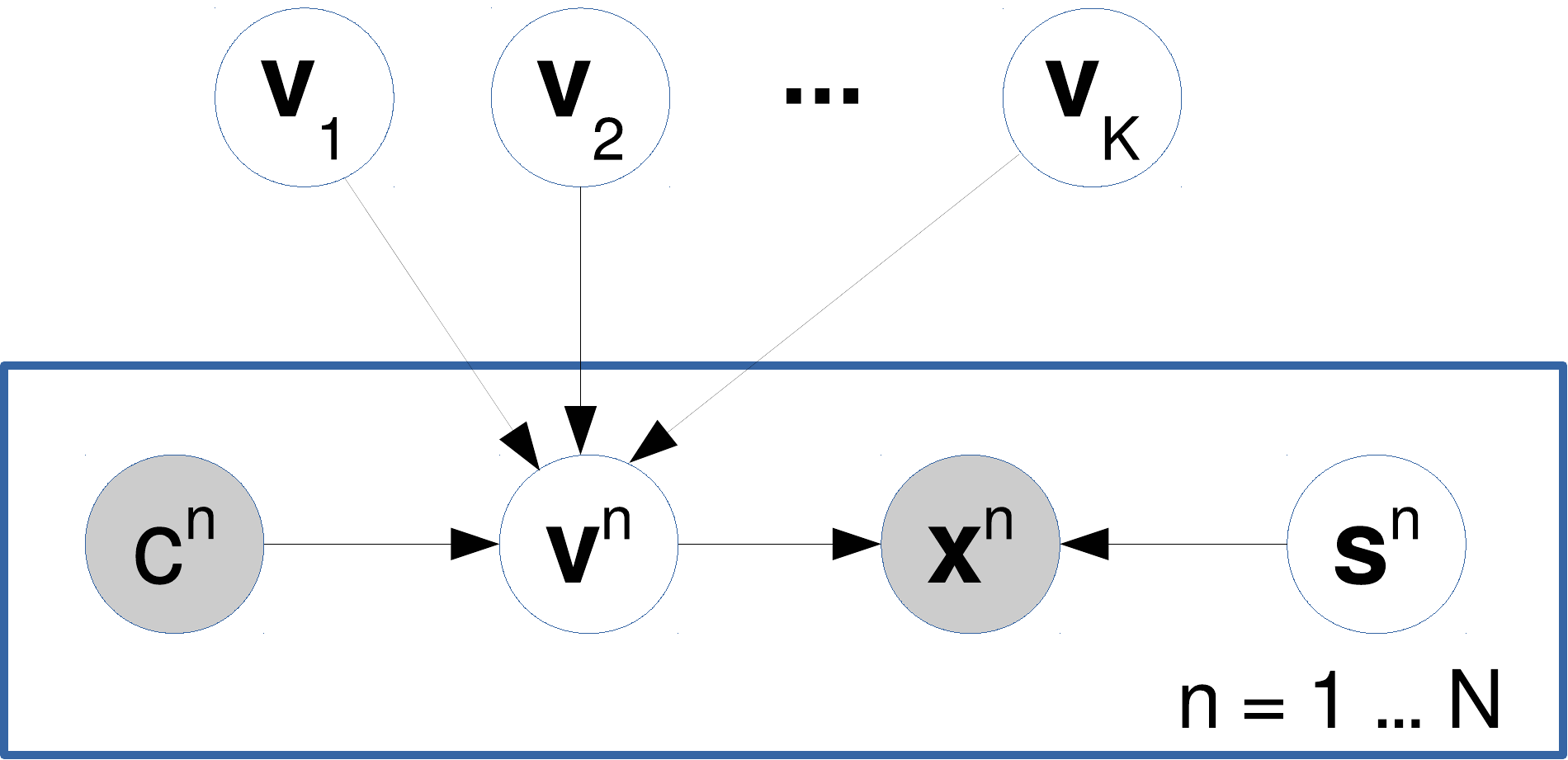} 
\end{center}
\vspace{-0.8em}
\caption{Graphical model representation for the proposed OL-VAE. 
Observed variables, the image/data instance ${\bf x}$ and the content label $c$, are shaded.
}
\vspace{-1.0em}
\label{fig:level_vae}
\end{figure}

\subsection{Unifying Graphical Model}\label{sec:principled}

Although the 
OL-VAE model in 
\autoref{fig:level_vae} is general enough to unify both OL-VAE and ML-VAE within the same framework, the main difference is the prior distribution of the content latents. We introduce a unified model to facilitate the expositions of how we can pose the prior distribution of the (ordinal) content latents (\autoref{sec:latent_priors}) and the variational inference (\autoref{sec:vi}).  

An observed data instance ${\bf x}$ is considered to be generated from a pair of latent vectors $({\bf v}, {\bf s})$, the content and the style. Since the content label $c \in \{1,\dots,K\}$ 
is available in data, it is reasonable to let ${\bf v}$ take one of $K$ predefined {\em content reference vectors} $\{{\bf v}_i\}_{i=1}^K$, where ${\bf v}_i$ is the 
content representation of $c=i$:
\begin{equation}
P({\bf v} \ | \ c=i, {\bf v}_1,\dots,{\bf v}_K) = \delta({\bf v}-{\bf v}_i) 
\label{eq:content_conditional}
\vspace{+0.3em}
\end{equation}
where $\delta(\cdot)$ is the Dirac's delta function. 
The reference vectors are random variables, and once sampled and fixed, ${\bf v}$ is associated {\em deterministically} to the content label $c$, its index. 

This differentiates ${\bf v}$ from ${\bf s}$ in that an individual random variable ${\bf s}$ exists for each instance ${\bf x}$, but there is one ${\bf v}_i$ that governs all instances ${\bf x}$ with the same content $c=i$. Using the plate notation, the graphical model can be defined as in \autoref{fig:level_vae}. Here $n$ indicates the data instance, among $N$. The full joint distribution can then be written as:
\begin{align}
&P\big( \{{\bf v}_i\}_{i=1}^K, 
  \{(c^n,{\bf v}^n,{\bf x}^n,{\bf s}^n)\}_{n=1}^N \big) = P({\bf v}_1,\dots,{\bf v}_K) \ \times \ \ \ \ \nonumber \\
&\ \ \ \prod_{n=1}^N 
  P(c^n) 
  P({\bf v}^n \ | \ c^n, \{{\bf v}_i\}_{i=1}^K) 
  P({\bf s}^n) P({\bf x}^n | {\bf v}^n, {\bf s}^n)
\label{eq:level_vae_full}
\end{align}
$P({\bf x}^n | {\bf v}^n, {\bf s}^n)$ is the decoder model that generates an image ${\bf x}$ from the pair of latents, and the choice of $P(c^n)$ is arbitrary. 


\begin{figure}
\begin{center}
\includegraphics[trim = 0mm 0mm 0mm 0mm, clip, scale=0.175
]{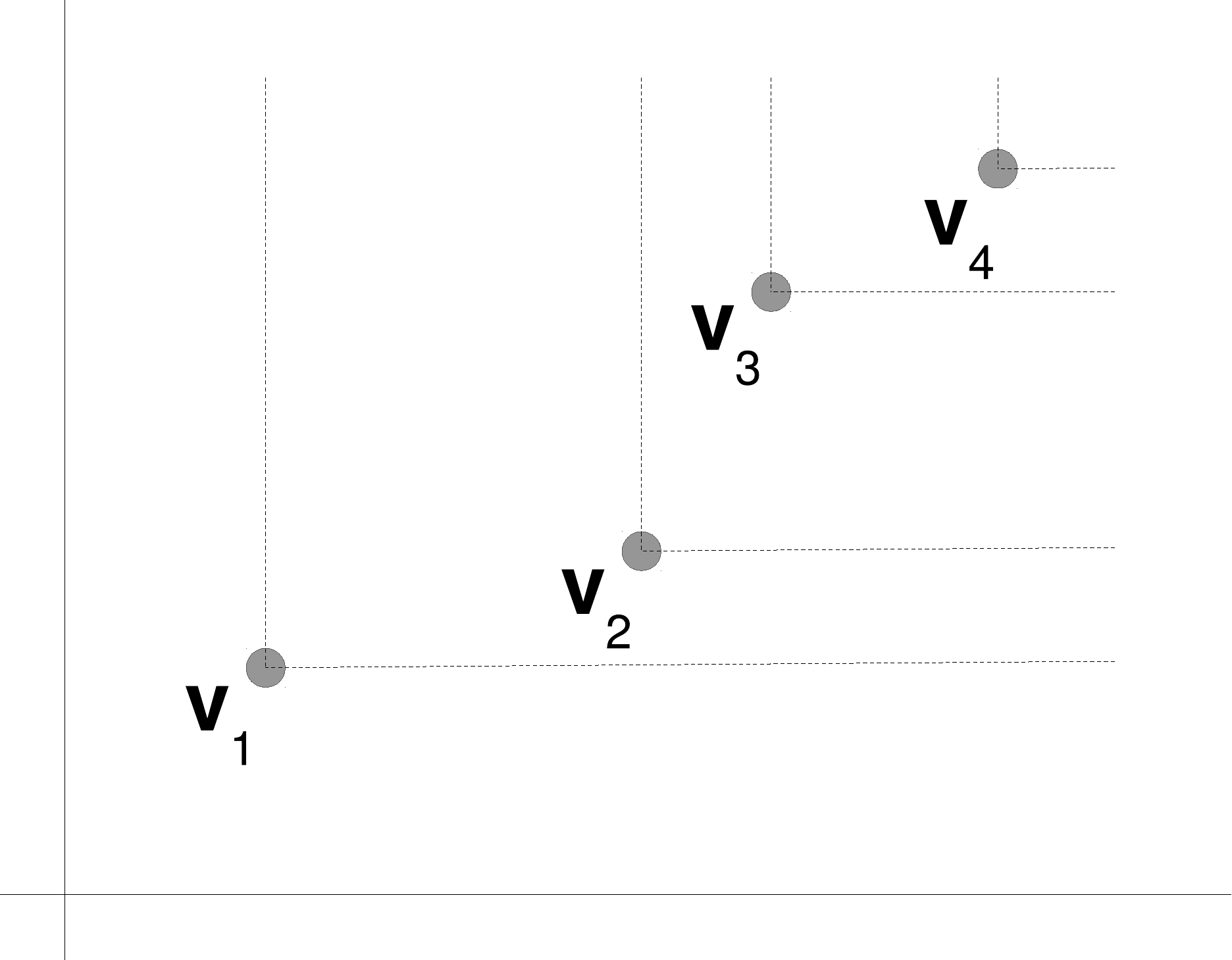} 
\includegraphics[trim = 0mm 0mm 0mm 0mm, clip, scale=0.175
]{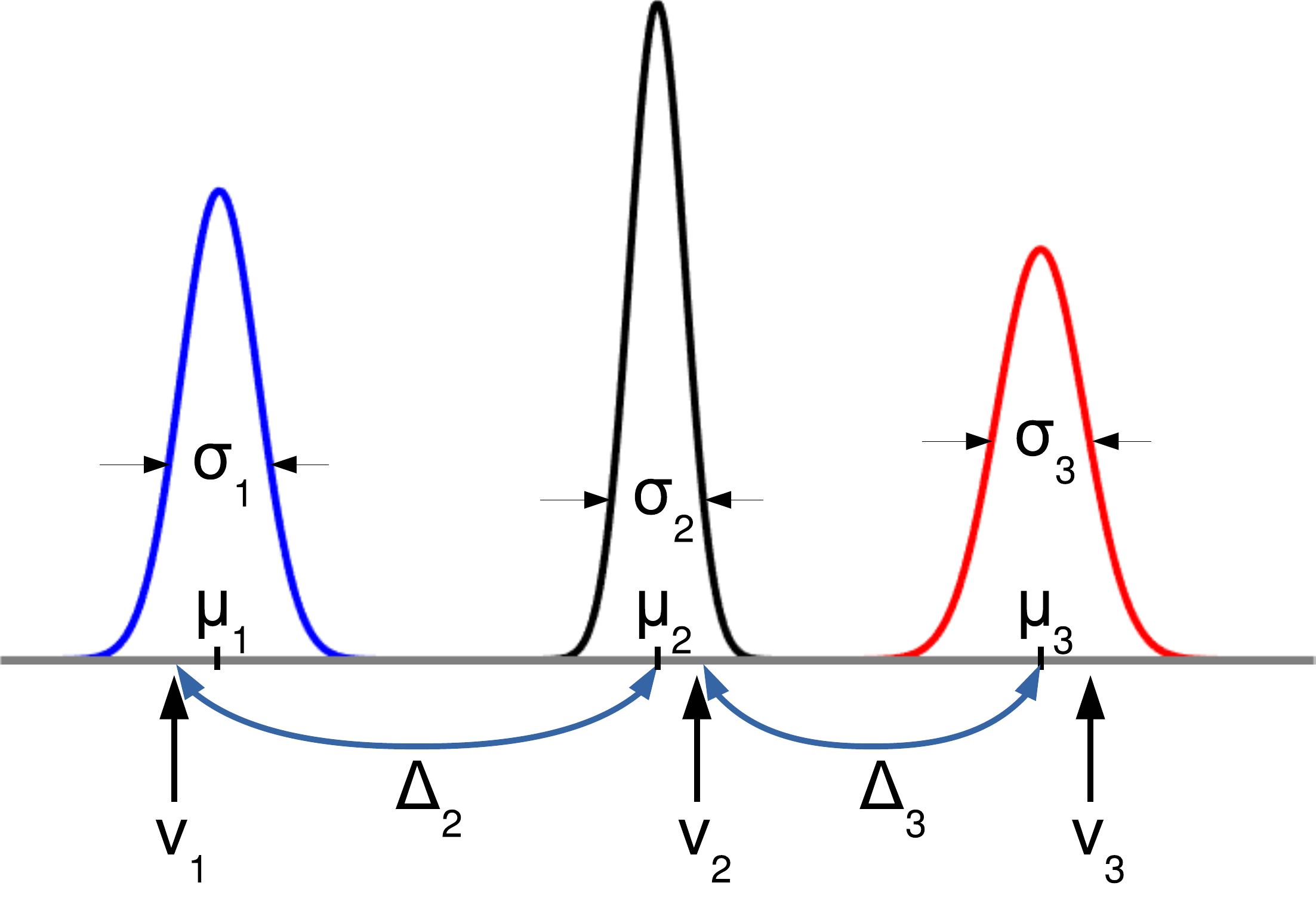} 
\end{center}
\vspace{-0.8em}
\caption{(\textbf{Left}) $K=4$ latent vectors (points) 
aligned in a poset formation in $\mathbb{R}^2$. For each vector ${\bf v}_i$, the adjoined horizontal and vertical lines specify the feasible quadrant where its superiors ${\bf v}_j$ for $j>i$ can be positioned. 
(\textbf{Right}) Conditional spacing model $P(v_1,v_2,v_3) = P(v_1) P(v_2|v_1) P(v_3|v_2)$ for $K=3$. Our choice of Gaussian conditional densities (\ref{eq:pv_olvae_conds_1}--\ref{eq:pv_olvae_conds_4}) together with the constraints \eqref{eq:pv_param_constrs} guarantees that the samples from $P(v_1,v_2,\dots,v_K)$ satisfy $v_1 \leq v_2 \leq \dots \leq v_K$ with high probability. 
}
\vspace{-0.8em}
\label{fig:poset_pvconds}
\end{figure}

\subsection{Latent Priors}\label{sec:latent_priors}

For the style prior, a natural choice is the iid standard normal, $P({\bf s}) = \mathcal{N}({\bf 0}, {\bf I})$ and $P(\{{\bf s}^n\}_{n=1}^N) = \prod_{n=1}^N \mathcal{N}({\bf s}^n; {\bf 0}, {\bf I})$, as in the VAE. For the content reference vectors, in the ML-VAE they used fully factorized standard normal distributions, $P({\bf v}_1,\dots,{\bf v}_K) = \prod_{i=1}^K \mathcal{N}({\bf v}_i; {\bf 0}, {\bf I})$. 
Using this prior in \eqref{eq:level_vae_full}, exactly recovers ML-VAE's full joint model. 
However, for the ordinal-valued content, where proximity in the content label values needs to be reflected in the latent space, the iid prior is suboptimal, unable to preserve the ordinal property. 


To have a more sensible ordinal content prior model, one would want to impose the triplet ordering constraint in \eqref{eq:content-distance-ineq}.  One way to meet this constraint is to have a {\em partially ordered set (poset)} for the content latents. For each dimension $l=1,\dots,d$, we make sure that the $l$-th elements of the latent vectors are ordered, namely $[{\bf v}_1]_l < [{\bf v}_2]_l < \cdots < [{\bf v}_K]_l$, where $[{\bf v}_i]_l$ is the $l$-th entry of the vector ${\bf v}_i$ (i.e., ${\bf v}_i = [ [{\bf v}_i]_1, [{\bf v}_i]_2, \dots, [{\bf v}_i]_d ]^\top$). Any set of $K$ vectors aligned in a poset formation satisfies the 
content-distance constraints \eqref{eq:content-distance-ineq}.  \autoref{fig:poset_pvconds} illustrates this for $d=2$-dim latent space. 
%
To impose the poset constraint in a prior, we propose a novel {\em conditional Gaussian spacing model}. 
It effectively places negligible probability density on configurations that violate the poset constraint. The model results in a joint Gaussian distribution over the $K$ vectors, fully correlated over $i=1,\dots,K$, yet fully factorized over dimensions $l=1,\dots,d$, making variational inference 
computationally tractable.

\textbf{Conditional Gaussian Spacing Model}. 
We consider a dimension-wise independent distribution, 
\begin{equation}
P({\bf v}_1,\dots,{\bf v}_K) = \prod_{l=1}^d P_l([{\bf v}_1]_l, \dots, [{\bf v}_K]_l),
\label{eq:pv_olvae_dimwise_indep}
\end{equation}
where $P_l$ is the density over $K$ variables from $l$-th dimension of the latent vectors. 
We model $P_l$ by a product of predecessor-conditioned Gaussians. 
For simplicity we drop the subscript $l$ in notation, and abuse $v_i$ to denote $[{\bf v}_i]_l$, $P(v_1,\dots,v_K)$ to refer to $P_l([{\bf v}_1]_l, \dots, [{\bf v}_K]_l)$. Now the model is:
\begin{equation}
P(v_1,\dots,v_K) = P(v_1) P(v_2|v_1) 
\cdots P(v_K|v_{K-1}), 
\label{eq:pv_olvae_conditioned}
\end{equation}
where the conditionals are defined as:
\begin{align}
P(v_1) &= \mathcal{N}(v_1; \mu_1, \sigma_1^2) \label{eq:pv_olvae_conds_1} \\ 
P(v_2|v_1) &= \mathcal{N}(v_2;  \underbrace{v_1+\Delta_2}_{:=\mu_2}, \sigma_2^2) 
\label{eq:pv_olvae_conds_2} \\
P(v_3|v_2) &= \mathcal{N}(v_3;  \underbrace{v_2+\Delta_3}_{:=\mu_3}, \sigma_3^2) 
\label{eq:pv_olvae_conds_3} \\ 
& \ \ \vdots \nonumber \\
P(v_K|v_{K-1}) &= \mathcal{N}(v_K;  \underbrace{v_{K-1}+\Delta_K}_{:=\mu_K}, \sigma_K^2)
\label{eq:pv_olvae_conds_4}
\end{align}
We define $P(v_i|v_{i-1})$ as a Gaussian centered at $\mu_i := v_{i-1} + \Delta_i$ with variance $\sigma_i^2$. That is, $\Delta_i$ ($> 0$) is the spread between the predecessor sample $v_{i-1}$ and the mean $\mu_i$ (See \autoref{fig:poset_pvconds} for the intuition). 
We consider $\{\sigma_i,\Delta_i\}_{i=1}^K$ and $\mu_1$ to be the free parameters of the model that can be learned from data. 
To guarantee that we meet the poset constraint, 
we make each conditional distribution (pillar) separated from its adjacent neighbors through the following constraints: 
\begin{equation}
\Delta_i \geq 3\sigma_i. 
\label{eq:pv_param_constrs}
\end{equation}
With \eqref{eq:pv_param_constrs}\footnote{
The inequality constraints \eqref{eq:pv_param_constrs} can be easily incorporated in the conventional unconstrained optimizer modules such as PyTorch and TensorFlow through trivial reparametrizations, e.g., $\sigma_i := \frac{\Delta_i}{3} \textrm{sigmoid}(\overline{\sigma}_i)$ and $\Delta_i := \exp(\overline{\Delta}_i)$, where $\overline{\Delta}_i$ and $\overline{\sigma}_i$ are the unconstrained optimization variables, and $\textrm{sigmoid}(x) = 1/(1+\exp(-x))$. 
Furthermore, we fix $\sigma_1=1$ to make the optimization numerically more stable.
}, the likelihood that $v_i \leq v_{i-1}$ is negligible
enforcing the desired ordering $v_1 < v_2 < \dots < v_K$. 


The joint density for (\ref{eq:pv_olvae_conds_1}--\ref{eq:pv_olvae_conds_4}) now admits a closed-form. 
Since everything is Gaussian and linear here, so the joint density $P(v_1,v_2,\dots,v_K)$ must be Gaussian. One can show that the means and covariances of the full joint Gaussian model can be written as follows (See Appendix~\ref{sec:joint_prior} for the proof):
\begin{align}
\mathbb{E}[v_i] &= \mu_1 + \Delta_2 + \cdots + \Delta_i \ \ (\textrm{for} \ \ i \geq 2) 
\label{eq:pv_joint_mean} \\
\mathbb{\textrm{Cov}}(v_i,v_j) &= \sigma_1^2 + \cdots +  \sigma_{\min(i,j)}^2 
\label{eq:pv_joint_cov}
\end{align}
E.g., for $K=3$, the joint distribution $P(v_1,v_2,v_3)$ is:
\begin{equation}
\mathcal{N}\Bigg( 
\underbrace{
  \begin{bmatrix}
    \mu_1 \\
    \mu_1+\Delta_2 \\
    \mu_1+\Delta_2+\Delta_3
  \end{bmatrix}
}_{:={\bf a}},
\underbrace{
  \begin{bmatrix}
    \sigma_1^2 & \sigma_1^2 & \sigma_1^2 \\
    \sigma_1^2 & \sigma_1^2 + \sigma_2^2 & \sigma_1^2 + \sigma_2^2 \\
    \sigma_1^2 & \sigma_1^2 + \sigma_2^2 & \sigma_1^2 + \sigma_2^2 + \sigma_3^2
  \end{bmatrix}
}_{:={\bf C}}
\Bigg)
\label{eq:pv_joint_ex}
\end{equation}
where we denote the mean vector and covariance matrix of the joint Gaussian by ${\bf a}$ and ${\bf C}$, respectively. 
Plugging this back in our original prior model \eqref{eq:pv_olvae_dimwise_indep}, 
we have:
\begin{equation}
P({\bf v}_1,\dots,{\bf v}_K) 
= \prod_{l=1}^d \mathcal{N}([{\bf V}]_l; {\bf a}_l, {\bf C}_l),
\label{eq:pv_olvae}
\end{equation}
where $[{\bf V}]_l := \big[ [{\bf v}_1]_l, \dots, [{\bf v}_K]_l \big]^\top$ is the $K$-dim vector collecting $l$-th dim elements from ${\bf v}_i$'s. Also, ${\bf a}_l$ and ${\bf C}_l$, for each $l=1,\dots,d$, are defined by (\ref{eq:pv_joint_mean}--\ref{eq:pv_joint_ex}) with their own free parameters, denoted as: $\big( \mu^{l}_{1}, \{\Delta^{l}_i, \sigma^{l}_i \}_{i=1}^K \big)$.
The covariance ${\bf C}_l$ are not diagonal. However, the model is factorized over $l=1,\dots,d$, the fact exploited in the next section to make the variational inference tractable.

\subsection{Variational Inference}\label{sec:vi}

Given the content-labeled data $\{({\bf x}^n, c^n)\}_{n=1}^N$, we approximate the posterior by the following variational density, decomposed into the content and style latents. The content posterior is further factorized over the content levels $c=1,\dots,K$,
\begin{equation}
Q\big( \{{\bf v}_i\}_{i=1}^K, \{{\bf s}^n\}_{n=1}^N \big) := \prod_{i=1}^K Q_c\big( {\bf v}_i | \{{\bf x}^n\}_{n \in G_i} \big) \prod_{n=1}^N Q_s\big( {\bf s}^n | {\bf x}^n \big),
\label{eq:vi_Q_form}
\end{equation}
where $G_i = \{n: c^n=i\}$ is the set of the training instances with content label $c=i$. For the encoders $Q_c$ and $Q_s$, we adopt deep networks that take an input ${\bf x}$ and output the means and variances of the Gaussian-distributed latents. However, since $Q_c$ requires a group of samples $\{{\bf x}^n\}_{n\in G_i}$ as its input, instead of adopting a complex group encoder such as the neural statisticians~\cite{nstatistician,homoencoder}, we use a simple product-of-expert rule, also adopted in ML-VAE:
\begin{equation}
Q_c\big( {\bf v} | \{{\bf x}^n\}_{n \in G} \big) \propto \prod_{n \in G}  Q_c({\bf v} | {\bf x}^n).
\label{eq:poe}
\end{equation}
Since each $Q_c({\bf v} | {\bf x}^n)$ is Gaussian, the product \eqref{eq:poe} admits a Gaussian (Appendix~\ref{sec:poe}), 
and 
the upper bound of the data log-likelihood 
can be written as (Appendix~\ref{sec:elbo}): 
\begin{align}
&\sum_{i=1}^K \mathbb{E}_{Q_c({\bf v}_i|G_i)} \sum_{n\in G_i} \mathbb{E}_{Q_s( {\bf s}^n | {\bf x}^n)} \Big[ \log P({\bf x}^n|{\bf v}_i,{\bf s}^n) \Big] \nonumber \\
& \ \ \ \ - \ \textrm{KL}\Bigg( 
  \prod_{i=1}^K Q_c({\bf v}_i | G_i) \bigg\Vert P({\bf v}_1,\dots,{\bf v}_K) 
\Bigg) \nonumber \\
& \ \ \ \ - \ \sum_{n=1}^N \textrm{KL}\Big( 
  Q_s\big( {\bf s}^n | {\bf x}^n \big) \big\Vert P({\bf s}^n) 
\Big),
\label{eq:elbo}
\end{align}
where $Q_c({\bf v}_i|G_i)$ is shortcut notation for $Q_c\big( {\bf v}_i | \{{\bf x}^n\}_{n \in G_i} \big)$. Here $P({\bf s}^n) = \mathcal{N}({\bf 0},{\bf I})$ as usual, and $P({\bf v}_1,\dots,{\bf v}_K)$ is our ordinal-constrained prior \eqref{eq:pv_olvae} derived in the previous section. 
The first reconstruction loss term and the third style KL divergence are the same as those of ML-VAE, with the key difference in the second term. Below, we make a full derivation for its closed-form formula.

\textbf{Content latent KL term}. Although both distributions in the KL term are Gaussians, the full dependency of ${\bf v}_i$'s over $i=1,\dots,K$ in $P({\bf v}_1,\dots,{\bf v}_K)$ can be problematic if $d$ and $K$ are large. Specifically, as the input dimensionality of the distribution is $d\cdot K$, the Cholesky decomposition of  $(d\cdot K \times d\cdot K)$ covariance matrix, required for computing the KL divergence, might be prohibitive if $d$ and/or $K$ are large. However, along the latent dimensions $l=1,\dots,d$, the prior distribution $P({\bf v}_1,\dots,{\bf v}_K)$ is factorized as in \eqref{eq:pv_olvae}. Thus for a dimension-wise factorized encoder model, $Q_c({\bf v}_i | G_i) = \prod_{l=1}^d Q_c([{\bf v}_i]_l | G_i)$, a standard choice in the VAE-based auto-encoding literature, we can reduce the complexity from $O((d \cdot K)^3)$ down to $O(d \cdot K^3)$. More formally, the second term in \eqref{eq:elbo} can be written as:
\begin{equation}
\sum_{l=1}^d \textrm{KL}\Bigg( 
  \prod_{i=1}^K Q_c([{\bf v}_i]_l | G_i) \bigg\Vert P_l([{\bf v}_1]_l,\dots,[{\bf v}_K]_l)
\Bigg).
\label{eq:content_kl}
\end{equation}
Each summand in \eqref{eq:content_kl} is a KL divergence between Gaussians and can be written (up to constant) as:
\begin{align}
\frac{1}{2} \Bigg( 
  \textrm{Tr}({\bf C}_l^{-1} {\bf S}_l) + ({\bf a}_l - {\bf m}_l)^\top {\bf C}_l^{-1} ({\bf a}_l - {\bf m}_l) 
+ \log \frac{|{\bf C}_l|}{|{\bf S}_l|}
\Bigg), 
\label{eq:content_kl_each}
\end{align}
where $\prod_{i=1}^K Q_c([{\bf v}_i]_l | G_i)$ is denoted by $\mathcal{N}({\bf m}_l, {\bf S}_l)$, with ${\bf S}_l$ diagonal by definition. The inverse and determinant of ${\bf C}_l$ can be computationally tractable as $K$ is typically not large\footnote{We utilize the \texttt{inverse()} and \texttt{cholesky()} functions in PyTorch 1.1, which also allow auto-differentiations.}.


\section{Related Work}\label{sec:related}

Learning succinct but meaningful and interpretable  representations of data is the main goal of (deep) representation learning. The learned representations can be not just useful for downstream tasks serving as features (e.g., higher-level classification), but also crucial for data analysis that often requires high quality of interpretability. Recent approaches to representation learning broadly fall into three types of learning setups. The first is the {\em supervised} setup~\cite{sup_reed,sup_yang,sup_kulkarni,sup_whitney,locatello_sup} that can make full use of factor-labeled data in discovering the underlying factors of variations, however, preparing a rich set of labeled data is often prohibitive in practice. The {\em unsupervised} setup~\cite{infogan16,aae16,beta_vae17,nica17,dip_vae18,factor_vae18,tcvae} aims to learn the factors solely from the observed data without any supervision. To make the  latent variables exclusively responsible for the variation of a unique aspect in the observed data, often referred to as the goal of {\em latent disentanglement}, most of the approaches augment the objective function of the VAE~\cite{vae14} with an additional regularization term that encourages factorization/independence of the prior induced from the variational encoder. However, it was recently proved that the unsupervised setup suffers from unidentifiability unless proper inductive bias or regularity conditions are imposed~\cite{impossibility}. 

The {\em semi-supervised} setup seeks to remedy the limitations of the previous two extremal setups. We particularly focus on recent work that aimed at modeling specific factors and separating them from the others, similar and closely related to ours. In~\cite{DBLP:journals/corr/abs-1901-08534}, the samples of the reference factor values (e.g., neutral expression for the facial emotion factor) are exploited as weak supervision to capture and learn the visual differences in the images due to the changes in the factor values. In~\cite{lecun_disent}, the weak supervision is given as partial labels of specific class categories, in which the goal is to make the factor with labels orthogonal to other latents. However, instead of directly modeling the latent vectors, one for each class value, they considered a deep image-to-factor network that takes an image as input and returns a latent vector as output, essentially performing single instance inference instead of the group inference our OL-VAE and ML-VAE adopt. In addition, to separate the labeled factor from others, they formed an objective function as a mix of the conditional GAN loss and the VAE loss, which potentially incurs the sensitivity issue in selecting the hyperparameter that trades off between the two loss terms.

Some recent works aim to deal with ordinal-valued labels in the VAE framework. However, their goals and setups are inherently different from ours. Among others, in~\cite{ordinal_vae}, they consider a setup where some ordinal paired data instances are available in the training data. More specifically, in addition to the unlabeled data instances,  
they have some pairs $\{({\bf x}^{(a)}, {\bf x}^{(b)})\}$ such that for a particular factor of interest (denoted by $f$), their factor values are ordered $f({\bf x}^{(a)}) > f({\bf x}^{(b)})$. 
In~\cite{ordinal_vae2}, they deal with an ordinal label problem setup while the idea is to introduce a variational posterior for the ordinal label, which is modeled as an ordinal regressor. The consequence is that, unlike our approach, they do not explicitly enforce the layout of the latent vectors to be aligned with the ordinal constraints.

\section{Experiments}\label{sec:expmts}

We demonstrate OL-VAE's 
capability of  separating the content factor from other sources of variation, namely 
style. We highlight the importance of modeling the ordinal structure of the content factor, a unique property of our OL-VAE, by comparing it to non-ordinal content-style disentangled models, specifically the ML-VAE~\cite{mlvae}. We describe below the 
datasets 
we use, detailing the specific choice of the content factor, how the labels are collected and defined.


\begin{itemize}
\item \textbf{Sprites dataset}~\cite{dsprites}. The dataset consists of binary images of sprites with variations in the shape (oval, square, and heart) and four geometric factors: scale (6 variation modes), rotation (40), and $X$, $Y$ translation (32 modes each). From the original dataset, we form two datasets to test the content-style disentanglement: 1) \texttt{Rotation-Sprites} takes the rotation as the content while the rest four factors (X/Y-pos, shape, and scale) as style, and 2) \texttt{Scale-Sprites} regards the scale as content and the rest as style. Both rotation and scale factors clearly entail ordinal semantics. For both datasets, we use $K=6$ ordinal levels by merging adjacent content groups together if needed. We also collect only the images of rotation angles between $0$ and $45$ degrees to avoid redundancy in the datasets. 
\item \textbf{IMDB-WIKI dataset}~\cite{Rothe-ICCVW-2015,Rothe-IJCV-2016}.  
The dataset contains face images of celebrities (actors/actresses) collected from the IMDB and Wikipedia websites. For each image, the age of the subject is annotated; it is calculated/estimated from the date when the photo was taken and the birth date of the subject. To have tightly cropped face images, we use the face detector outputs comprised of both detection confidence scores and the bounding boxes that are provided in the dataset. We discarded images containing multiple faces, which was done by collecting only those images with high first face detection score and zero second detection score (not detected). After further removing faulty images contained in the original dataset, we ended up with 224,418 cropped face images (184,363 from IMDB and 40,055 from Wikipedia), out of the original 523,051 images.  
We used the image size $(64 \times 64)$ pixels, and set IMDB as the training set and Wikipedia as the test set.
We let the age serve as the content factor. To this end, we partition age values ranging from $5$ to $90$ into $17$ age groups. Each group covers non-overlapped 5-year segments. The average age group index is $5.89 \pm 2.62$ (about $34$ years old) for both Wikipedia and IMDB.  All other factors of variation, including subject ID and head pose, are considered as style (unlabeled). The data is highly noisy and biased, including a mix of gray/color images, cartoon images, and even stamp portraits. 
\item \textbf{Synthetic Pizza Calorie Dataset}. We use the dataset  in~\cite{papadopoulos19cvpr}, which contains clip-art style images of pizza. The images are synthetically generated by controlling the ingredients to be placed (ten ingredients, e.g., pepperoni, bacon, black olive, and basil) as well as the view angle, background, and the position of pizza in the image. There are 5,468 images, with $90\%$ / $10\%$ split for train/test using the original protocol. We used the image size $(64 \times 64)$ pixels. 
We add the {\em calorie} content of the pizza to~\cite{papadopoulos19cvpr} by computing the calories, albeit not very precisely, using look-up from a standard nutrition table\footnote{We used the nutrition tables in https://modpizza.com/nutrition/, which contain nutrition facts specifically tailored to pizzas. We assume standard ingredient amounts.} for known pizza ingredients. After computing the calorie values, we discretized them into five groups by roughly equal size binning: lowest, low, medium, high, and highest calorie cohorts. 
%
%
%
\end{itemize}
\begin{figure}[t!]
\vspace{-2.0em}
\begin{center}
\begin{subfigure}{0.5\textwidth}
\centering
\includegraphics[trim = 8mm 0mm 15mm 0mm, clip, scale=0.335
]{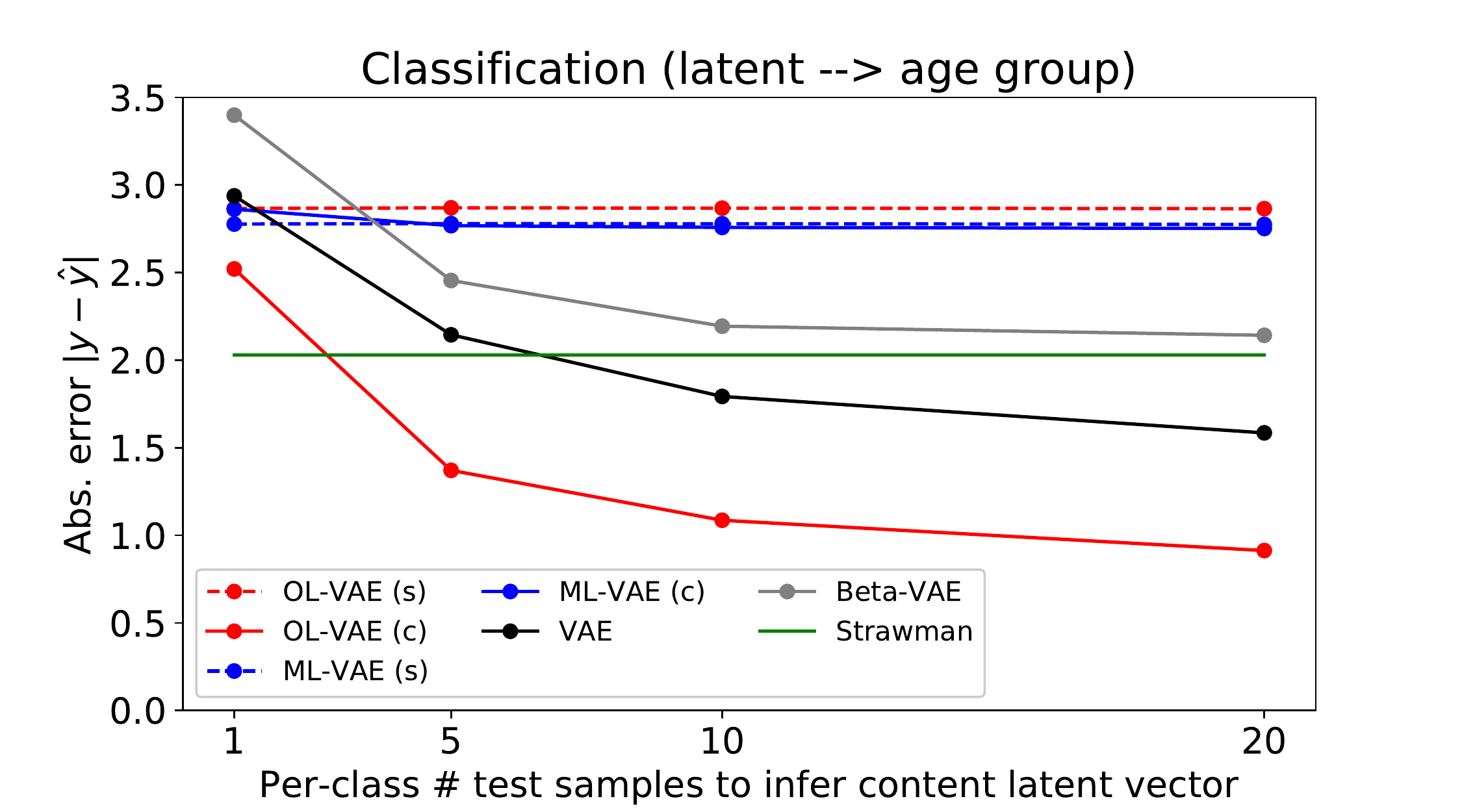}
\caption{Classification on IMDB-WIKI}
\end{subfigure}
\\
\begin{subfigure}{0.5\textwidth}
\centering
\includegraphics[trim = 10mm 0mm 13mm 0mm, clip, scale=0.335
]{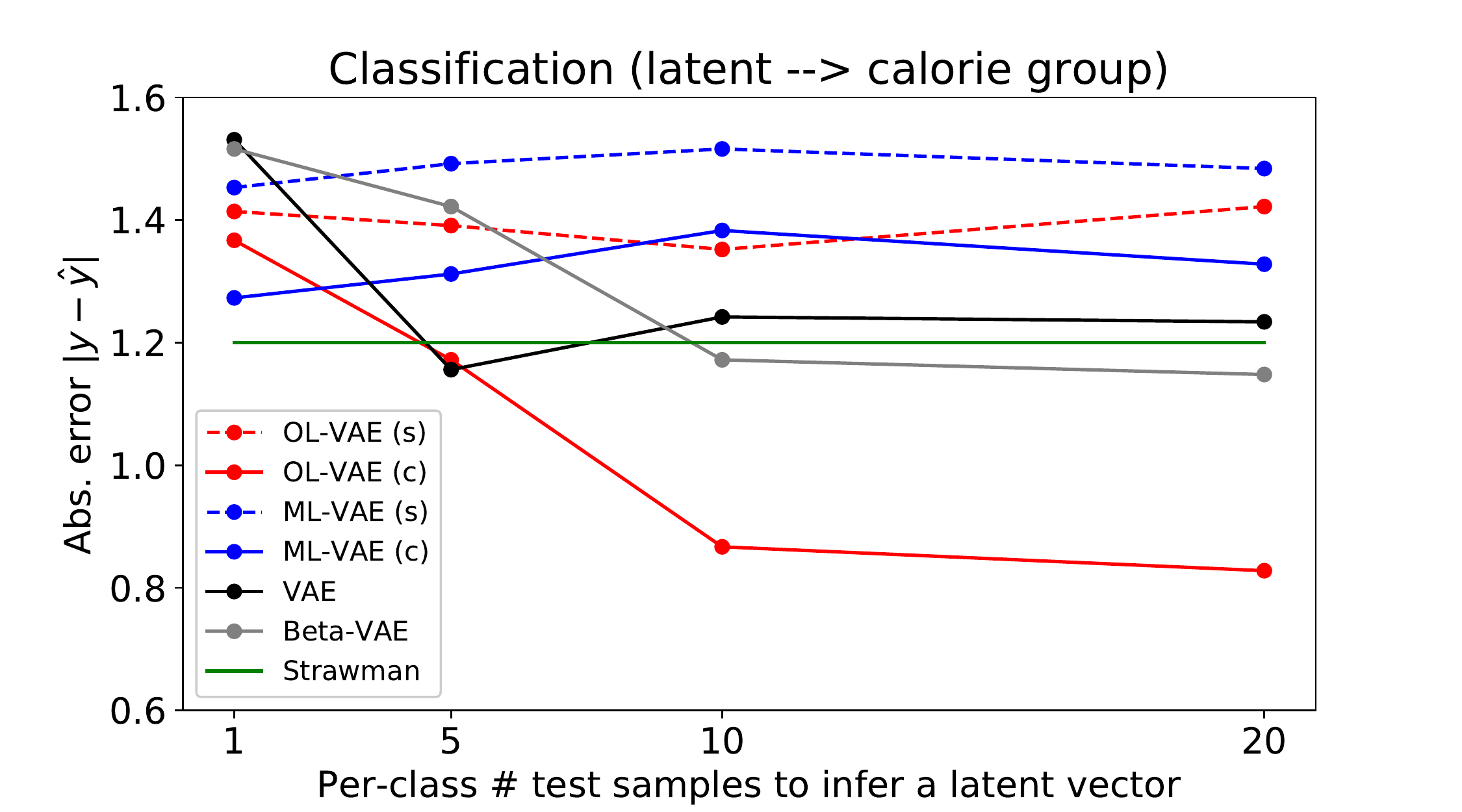}\caption{Classification on Pizza}
\end{subfigure}
\end{center}
\vspace{-0.5em}
\caption{Classification results on IMDB-WIKI and Pizza datasets. For OL-VAE and ML-VAE, $s$ and $c$ in the parenthese indicate that style and content latent vectors are used as classification covariates, respectively. Age and calorie variables indicate respective group indexes (thus, error of $1$ on e.g., IMDB-WIKI represents $5$ years.)
}
\vspace{-1.0em}
\label{fig:classify_embeds}
\end{figure}


\begin{figure}[t!]
\vspace{-2.0em}
\centering
\begin{subfigure}{0.5\textwidth}
\centering
\includegraphics[trim = 5mm 0mm 17mm 0mm, clip, scale=0.335
]{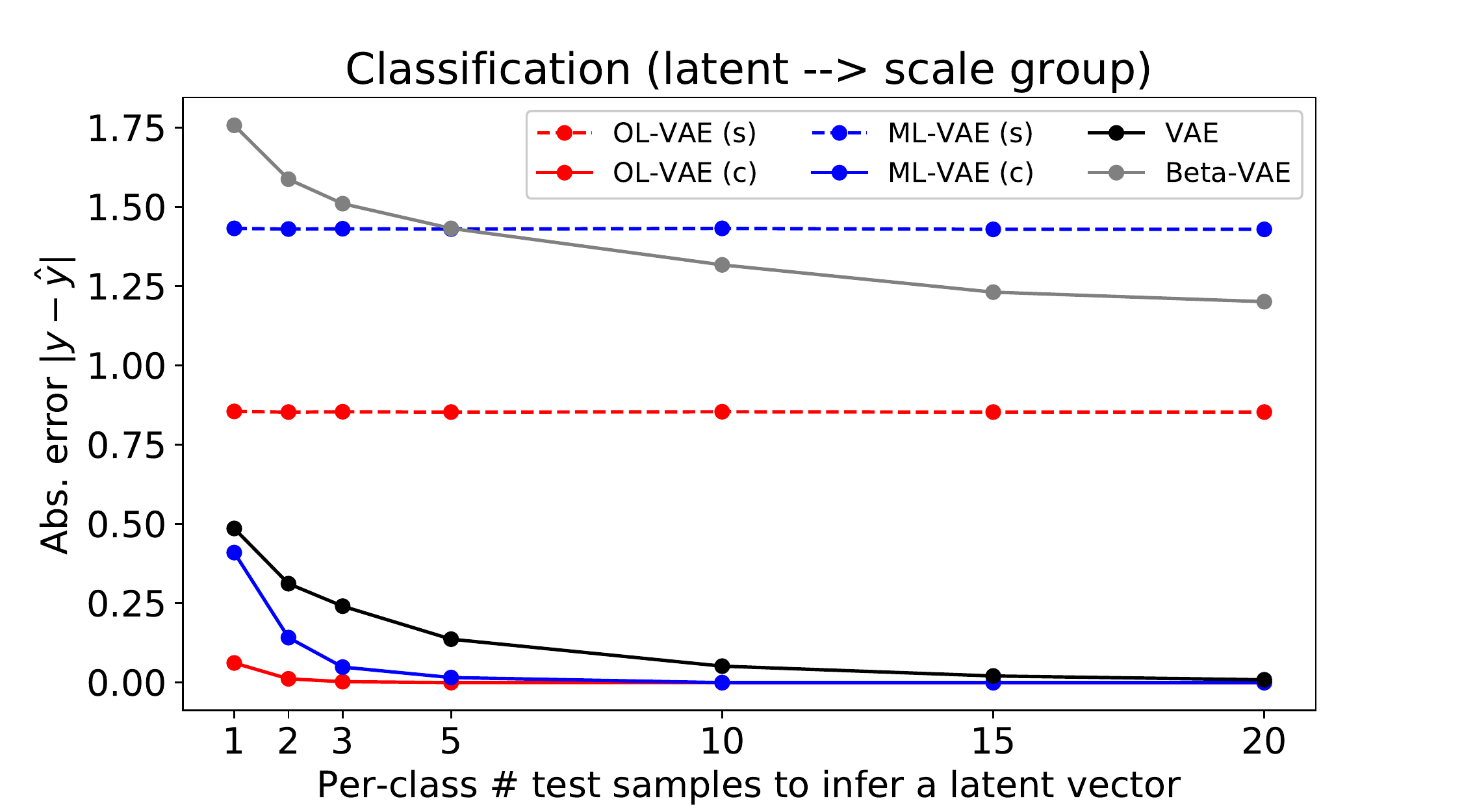}
\includegraphics[trim = 2mm 0mm 17mm 0mm, clip, scale=0.335
]{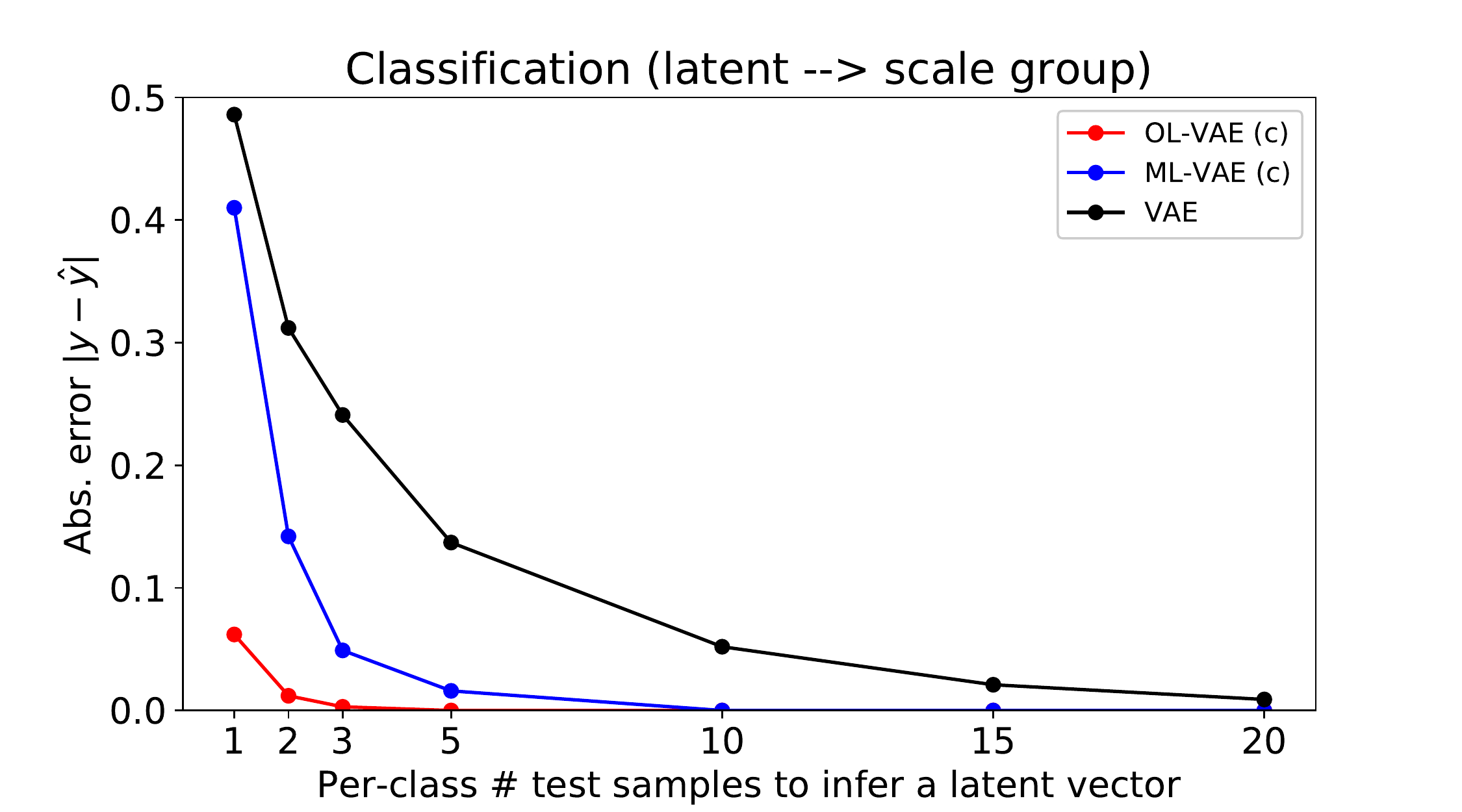} 
\vspace{-0.3em}
\caption{Scale-Sprites
}
\end{subfigure}
\begin{subfigure}{0.5\textwidth}
\centering
\includegraphics[trim = 5mm 0mm 17mm 0mm, clip, scale=0.335
]{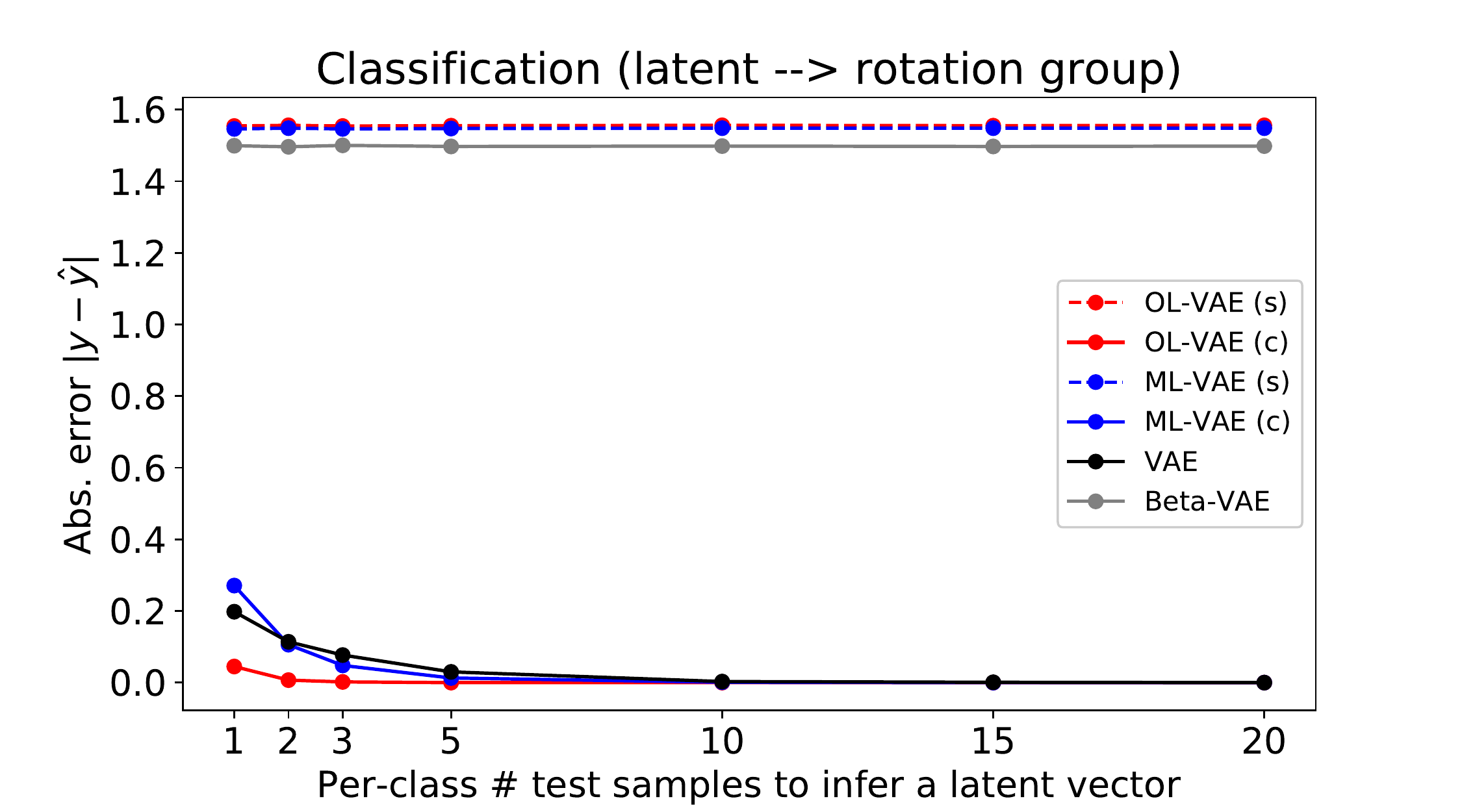}
\includegraphics[trim = 2mm 0mm 17mm 0mm, clip, scale=0.335
]{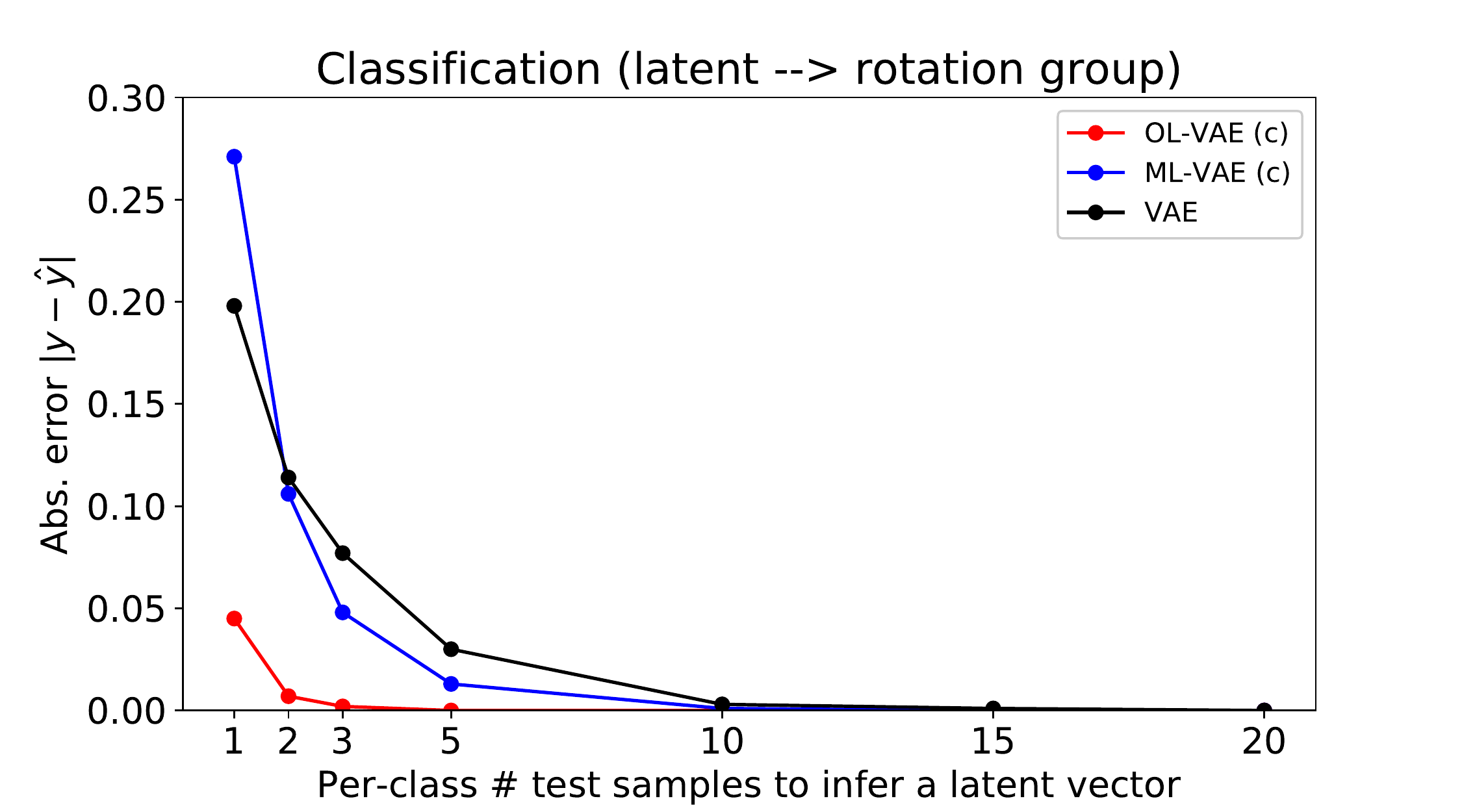} 
\vspace{-0.3em}
\caption{Rotation-Sprites
}
\end{subfigure}
\vspace{-0.3em}
\caption{Classification results on (a) Scale-Sprites and (b) Rotation-Sprites datasets. For each, 
(Top) shows classification errors for all models, and 
(Bottom) highlights the best three models. 
}
\label{fig:classify_sprites}
\vspace{-1.5em}
\end{figure}

\subsection{Competing Models}\label{sec:baselines}


In the empirical study, we focus on demonstrating the utility of ordinal content prior modeling facilitated by our OL-VAE model, highlighting the improvement over the non-ordinal content-style disentangled approaches, namely ML-VAE~\cite{mlvae}. To this end, we make all the setups including the model architectures equal for both models, but the prior modeling part. 
In particular, for the encoder/decoder architectures, we adopt models similar to those in ML-VAE: compositions of four/five convolutional or transposed convolutional layers and two/three fully connected layers, whereas for the decoder model $P({\bf x}|{\bf v},{\bf s})$, we set the output distribution to be Bernoulli; the normalized pixel value $x \in [0,1]$ is the mean of a Bernoulli process.

For baselines, we also compare with the VAE model~\cite{vae14}, and the disentanglement encouraging $\beta$-VAE~\cite{beta_vae17}. These are unsupervised models, unable to utilize the content labels during the model training. For these models, we use similar encoder/decoder architectures, but for fair comparison with the models with style and content latent vectors, the latent dimensions 
are set to $\textrm{dim}({\bf s}) + \textrm{dim}({\bf v})$, the sum of the style and content dimensions of OL-VAE. More specifically, the latent dimensions are: $\textrm{dim}({\bf s})=\textrm{dim}({\bf v})=50$ for IMDB-WIKI and Pizza datasets, and $\textrm{dim}({\bf s})=\textrm{dim}({\bf v})=10$ for the sprites datasets. 
The batch size is $256$ for all datasets.
%



\begin{figure}[t!]
\vspace{-2.0em}
\centering
\begin{subfigure}{0.145\textwidth}
\centering
\includegraphics[trim = 5mm 
8mm 8mm 8mm, clip, scale=0.114
]{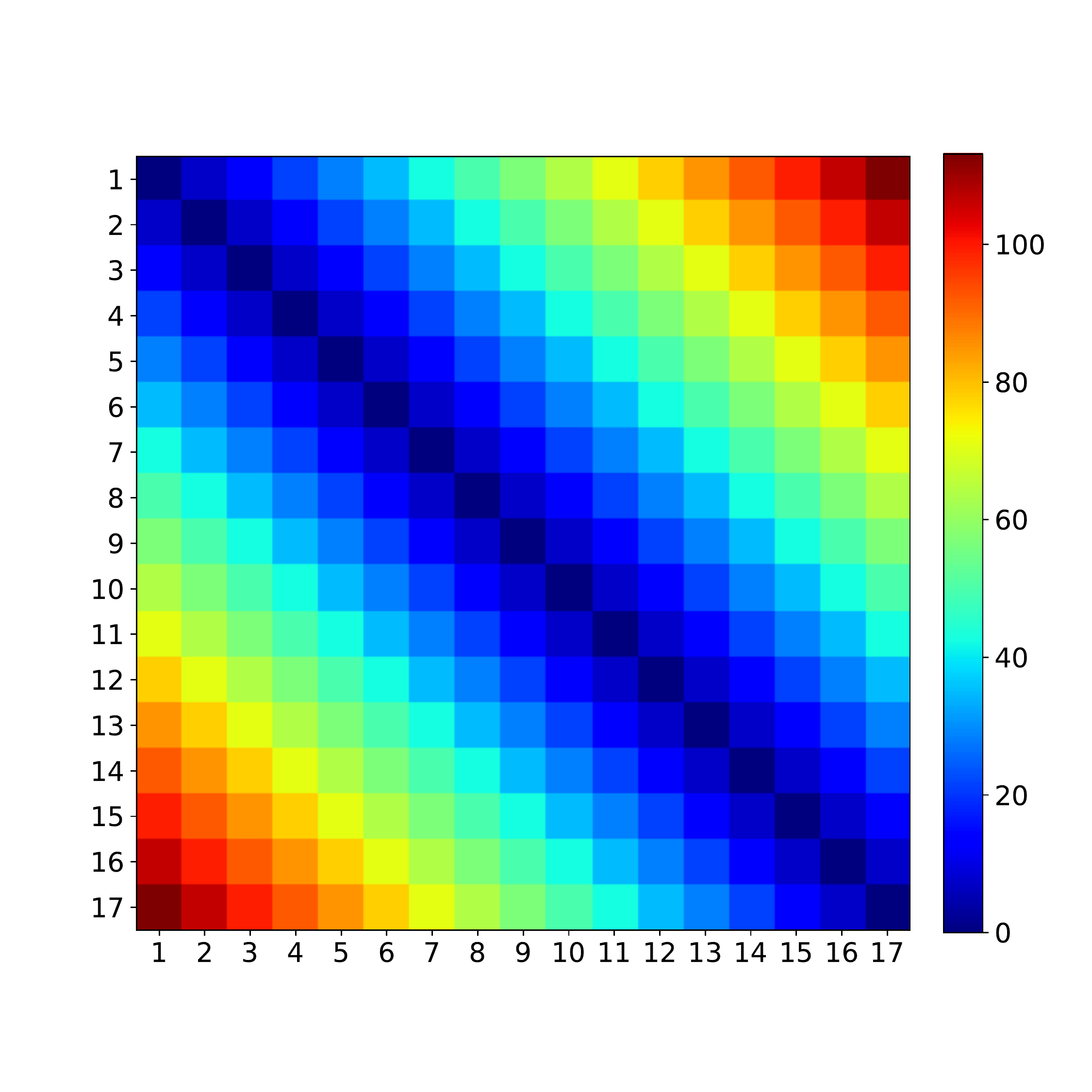} 
\vspace{-1.1em}
\caption{An ideal map}
\end{subfigure} 
\begin{subfigure}{0.145\textwidth}
\centering
\includegraphics[trim = 8mm 8mm 8mm 8mm, clip, scale=0.114
]{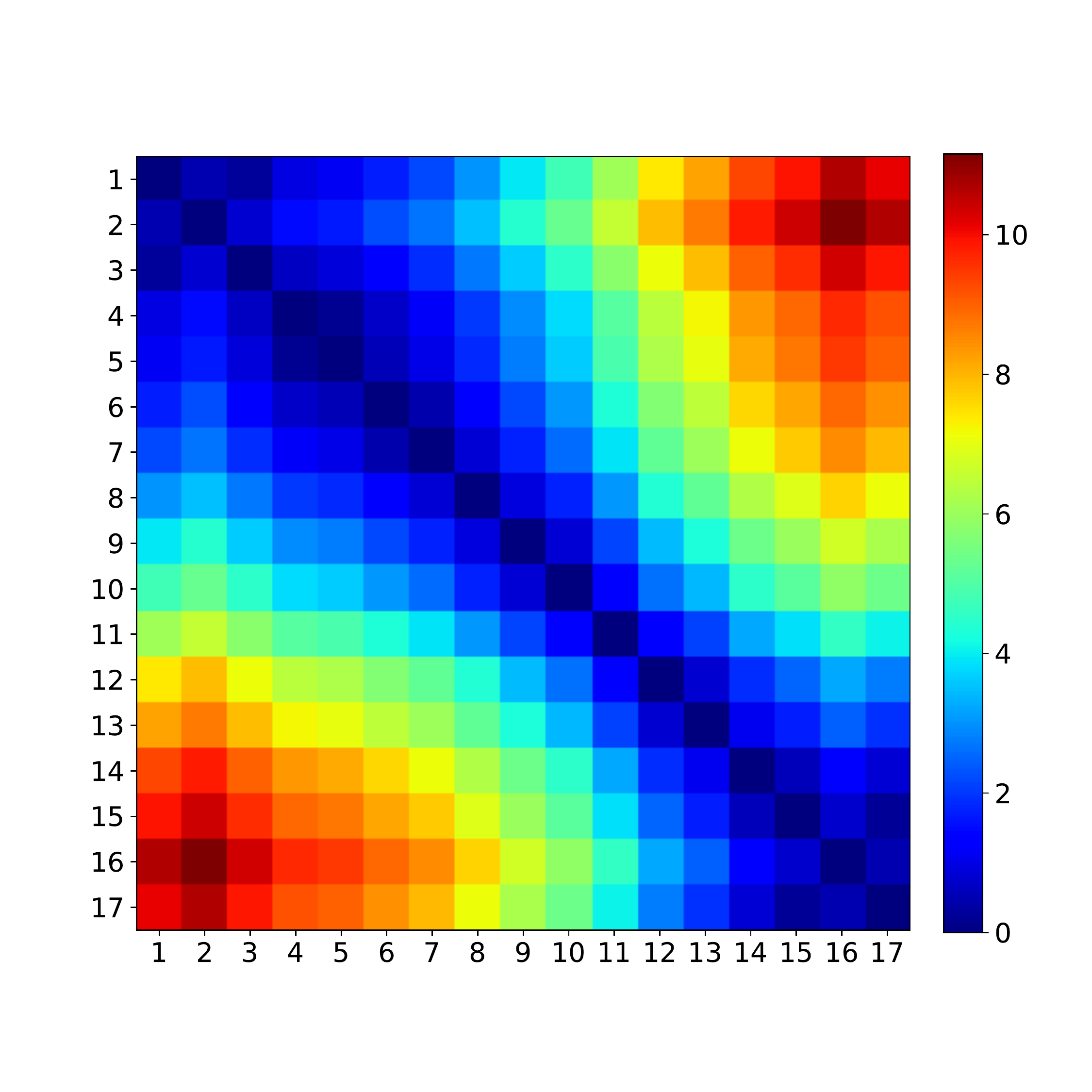} 
\vspace{-1.1em}
\caption{OL-VAE}
\end{subfigure} 
\begin{subfigure}{0.145\textwidth}
\centering
\includegraphics[trim = 8mm 8mm 8mm 8mm, clip, scale=0.114
]{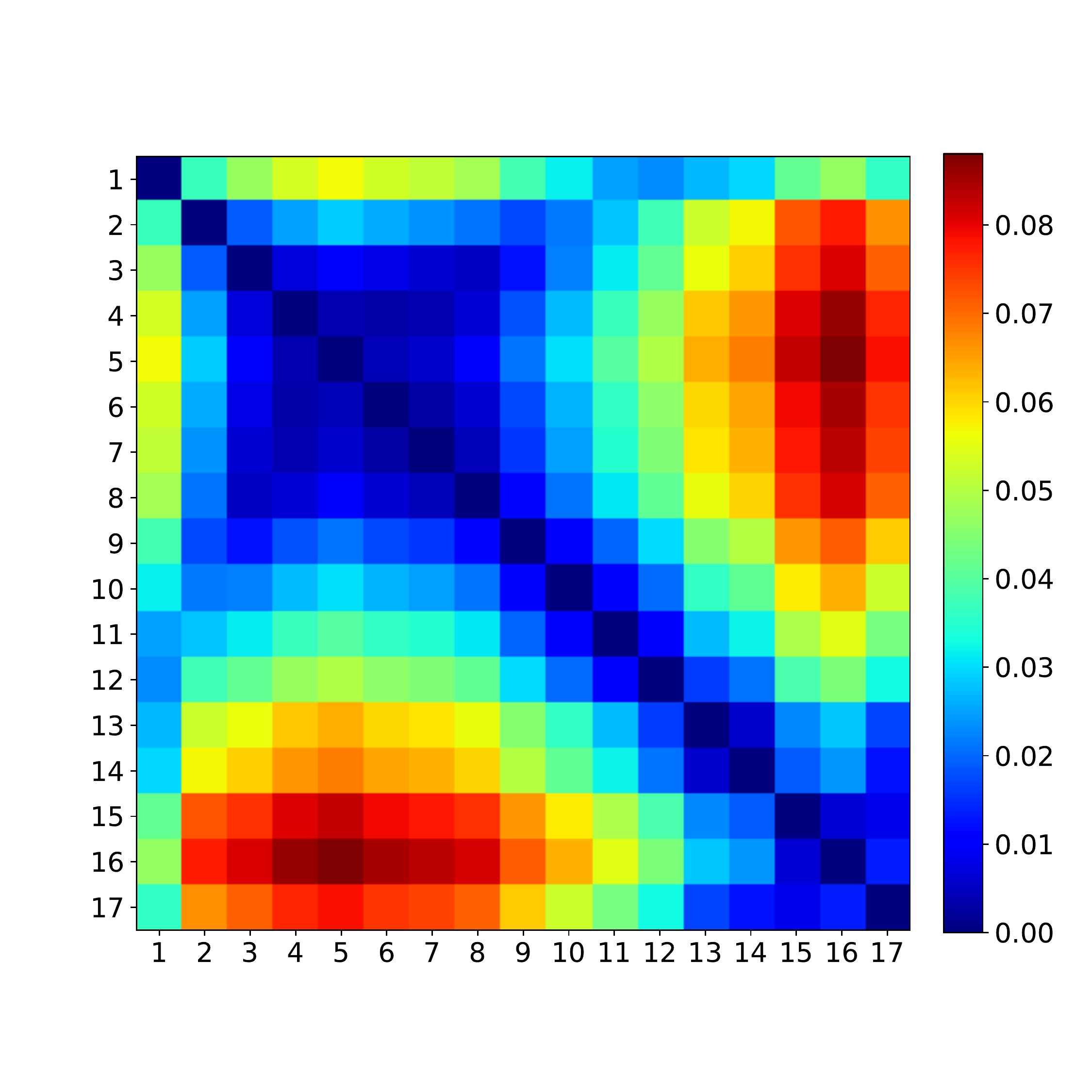} 
\vspace{-1.1em}
\caption{ML-VAE}
\end{subfigure}
\centering
\caption{(IMDB-WIKI dataset) The distance maps of the learned content latent vectors for (b) our OL-VAE and (c) ML-VAE.  Each $(i,j)$ entry of the $(K \times K)$ map with $K=17$ encodes the distance between two content vectors, $||{\bf v}_i-{\bf v}_j||$. Gradual and smooth increase of the distance from diagonals to off-diagonals, which looks more evident in our OL-VAE, indicates good alignment between the ordinal content labels and the positions of the latent vectors. As a reference, in (a) we also visualize {\em an} ideal distance map where the latent vectors are equally spaced, that is, ${\bf v}_i = {\bf v}_{i-1} + {\boldsymbol\delta}$ for all $i=2,\dots,K$, with some constant vector ${\boldsymbol\delta}$.
}
\label{fig:embed_dist_imdbwiki}
\vspace{-1.5em}
\end{figure}

\subsection{Latent to Content Prediction 
(Quantitative)}\label{sec:classification}

For a well-trained model, we anticipate the content latent ${\bf v}$ only contains the content information of ${\bf x}$, while the style latent vector ${\bf s}$ should ideally not entail content, instead capturing style. To test this capability, we form a classification problem: for each paired data $({\bf x},c)$, we encode ${\bf x}$ to have its  style/content latent representation, that is, ${\bf s}\sim Q_s(\cdot|{\bf x})$ and ${\bf v} \sim Q_c(\cdot|{\bf x})$, then either ${\bf s}$ or ${\bf v}$ serve as input covariates for classification toward the target content class label $c$. Note that ${\bf v}$ should be highly predictive of $c$, while ${\bf s}$ should not, leading to chance level classifier performance.

To form the content vector ${\bf v}$, we use $20$ samples from the same content group (i.e., ${\bf x}$'s with the same label $c$) for a group inference as in \eqref{eq:poe}. The classifier is trained with the IMDB dataset in the IMDB-WIKI and the preset training split for the Pizza, and we test on the WIKI in the IMDB-WIKI and the preset test split for the Pizza. At test time, we infer the covariates ${\bf v}$ using $M$ test samples per content class $c$ via the product of experts rule. We vary $M$ from $1$ to $20$. This setup is applied to OL-VAE and ML-VAE in the same manner. For the baseline VAE and $\beta$-VAE, which only have a single latent vector ${\bf z}$, we similarly accumulate $M$ group samples at the test time to reinforce the latent inference (i.e., the instance-wise variational posteriors $Q({\bf z}|{\bf x})$ are accumulated via the product of experts rule to have the covariates ${\bf z}$). As the target label $c$ has ordinal scale, we report the absolute error $|c-\hat{c}|$, instead of the $0/1$  loss.

The results are shown in Fig.~\ref{fig:classify_embeds} for the  IMDB-WIKI and Pizza datasets, and 
Fig.~\ref{fig:classify_sprites} for Rotation- and Scale-Sprites. 
For our OL-VAE, prediction using the style latent vector is significantly worse than that when using the content latent, indicating the style vector ${\bf s}$ carries little information about the content. 
For example, on IMDB-WIKI the mean abs error (MAE) of 
the mean age index guess of $5.89$ (a strawman predictor) is $2.03$,
indicating that ${\bf s}$ is not suggestive of age. Similarly, for the Pizza dataset, the mean guess of $2$ would yield the MAE of $1.2$, suggesting the lack of dependence between $s$ and the calorie content.
More importantly, our OL-VAE, throughout all datasets, achieves the lowest classification error (as $M$ increases) for predicting content using the content latent vector. 
The classification error is dramatically reduced as we increase the number of samples $M$ in the content group. This means that more evidence is helpful for inferring a correct content latent vector. On the other hand, the performance of ML-VAE trails that of OL-VAE and the strawman alike, often unaffected by the sample size $M$ (e.g., the Pizza dataset), implying that the learned content latent is not salient enough to predict the content value. In summary, these results signify that for high separation between the style and content factors, it is crucial to both preserve the ordinal structure in the latent embedding and align the content vectors with the content labels, as done by our OL-VAE model. 

For the baseline VAE and $\beta$-VAE, increasing the sample size lowers the classification error.  However, as there is mixing of content-style in the latent ${\bf z}$, predictive value of ${\bf z}$ is lower than what the isolated content latent could offer (in OL-VAE or ML-VAE); this is reflected in higher content prediction error for those unsupervised models.
\subsection{Visualization of Learned Content Vectors 
}\label{sec:learned}

To verify our claim that the learned content latent vectors are well aligned with the ordinal labels, we visualize the distance map of pairwise distances $||{\bf v}_i-{\bf v}_j||$ for $i,j=1,\dots,K$.
Specifically, for the test sample instances from each content group $c=i$, we infer ${\bf v}_i$'s for ML-VAE and OL-VAE using $Q_c({\bf v} | {\bf x}_{G_i})$.  We then compute the pairwise distances to form a $(K \times K)$ distance map. The maps are shown in Fig.~\ref{fig:embed_dist_imdbwiki} for the IMDB-WIKI dataset. For OL-VAE, we see the expected low values of distances on the diagonal and increasingly high values in the corners (mimicking a desired absolute loss $|c_i - c_j|$, indicated in the ideal map in the left-most image). This pattern is less pronounced for the ML-VAE, which exhibits a deformed heat map, indicating poor alignment.

\subsection{Swapping Content Values (Qualitative)}\label{sec:swapping}

Next, we demonstrate the content swapping capability of our OL-VAE model. More specifically we seek to qualitatively control the content aspect of an image by synthesizing new images of different content values while preserving the original style. This capability is only possible when the model achieves high degree of separation between style and content latents as in our OL-VAE. 

Suppose we use two reference images ${\bf x}^\textrm{st}$ and ${\bf x}^\textrm{co}$. We encode each of them to obtain their style and content embeddings. More formally,
\begin{align}
& {\bf s}^\textrm{st} \sim Q_s({\bf s}|{\bf x}^\textrm{st}), \ \ \ \ 
{\bf v}^\textrm{st} \sim Q_c({\bf v}|{\bf x}^\textrm{st}) \\
& {\bf s}^\textrm{co} \sim Q_s({\bf s}|{\bf x}^\textrm{co}), \ \ \ 
{\bf v}^\textrm{co} \sim Q_c({\bf v}|{\bf x}^\textrm{co})
\end{align}
As we have a learned decoder model, we can reconstruct the original image using the corresponding pair of style and content latent vectors, namely $P({\bf x}|{\bf v}^\textrm{st}, {\bf s}^\textrm{st})$. 
Extending the idea, one can consider {\em swapping} of content latents as a task of hallucinating a new image using the style vector ${\bf s}$ from one source and the content vector ${\bf v}$ from another. For instance, we can 
generate a new image that possesses the style of ${\bf x}^\textrm{st}$ and the content of ${\bf x}^\textrm{co}$:
\begin{equation}
{\bf x}^{\textrm{co}+\textrm{st}} \sim P({\bf x}|{\bf v}^\textrm{co}, {\bf s}^\textrm{st}).
\end{equation}
%
The better the model separates the content from the style, the more visually clear the chosen style and content will be preserved in the hallucinated image. 

\begin{figure*}
\vspace{-2.0em}
\begin{center}
\centering
\begin{subfigure}[t]{1.0\textwidth}
\centering
\includegraphics[trim = 0mm 0mm 0mm 0mm, clip, scale=0.52
]{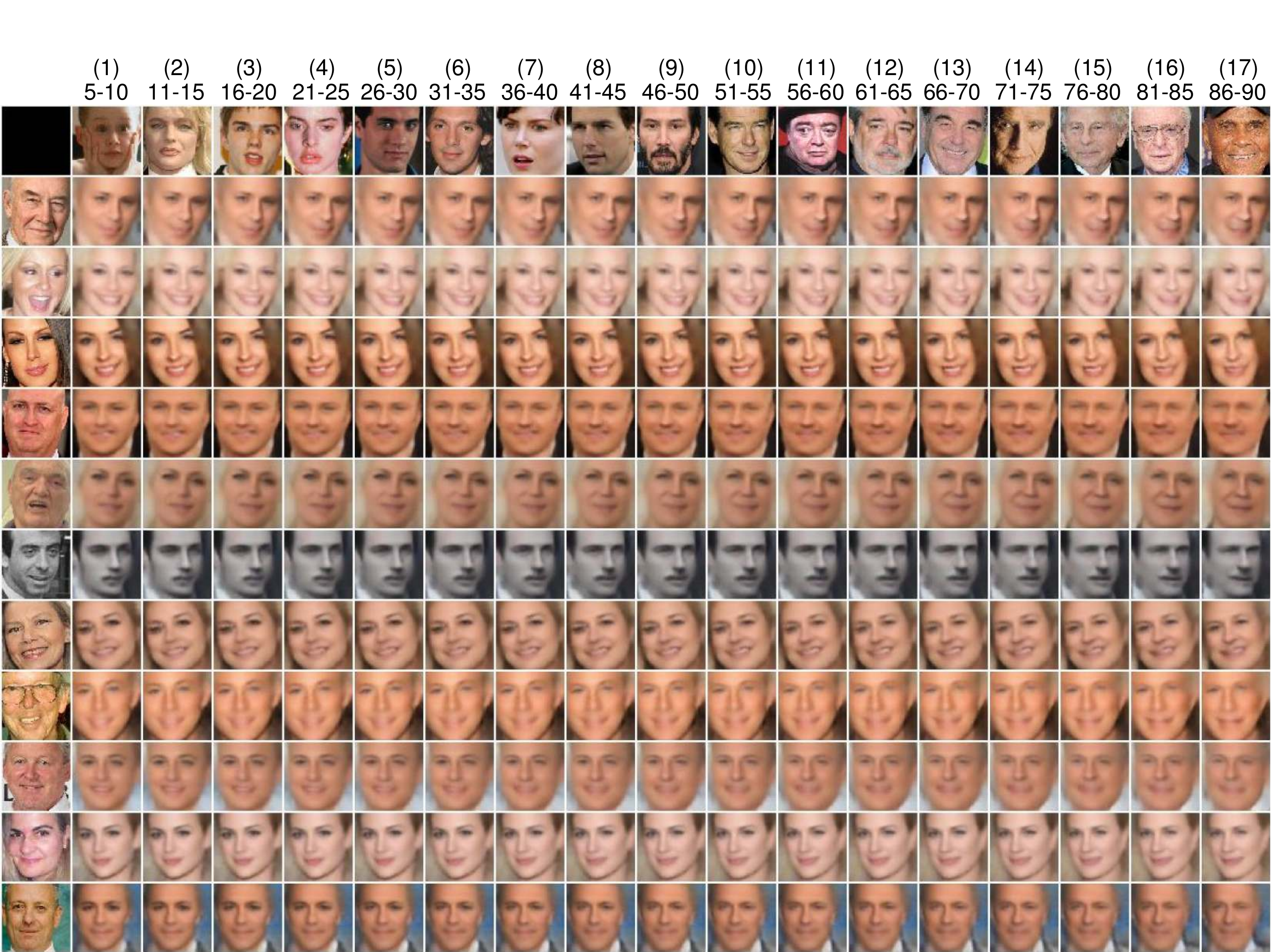}
\end{subfigure} \\ \vspace{+1.0em}
\begin{subfigure}[t]{1.0\textwidth}
\centering
\includegraphics[trim = 0mm 0mm 0mm 0mm, clip, scale=0.510
]{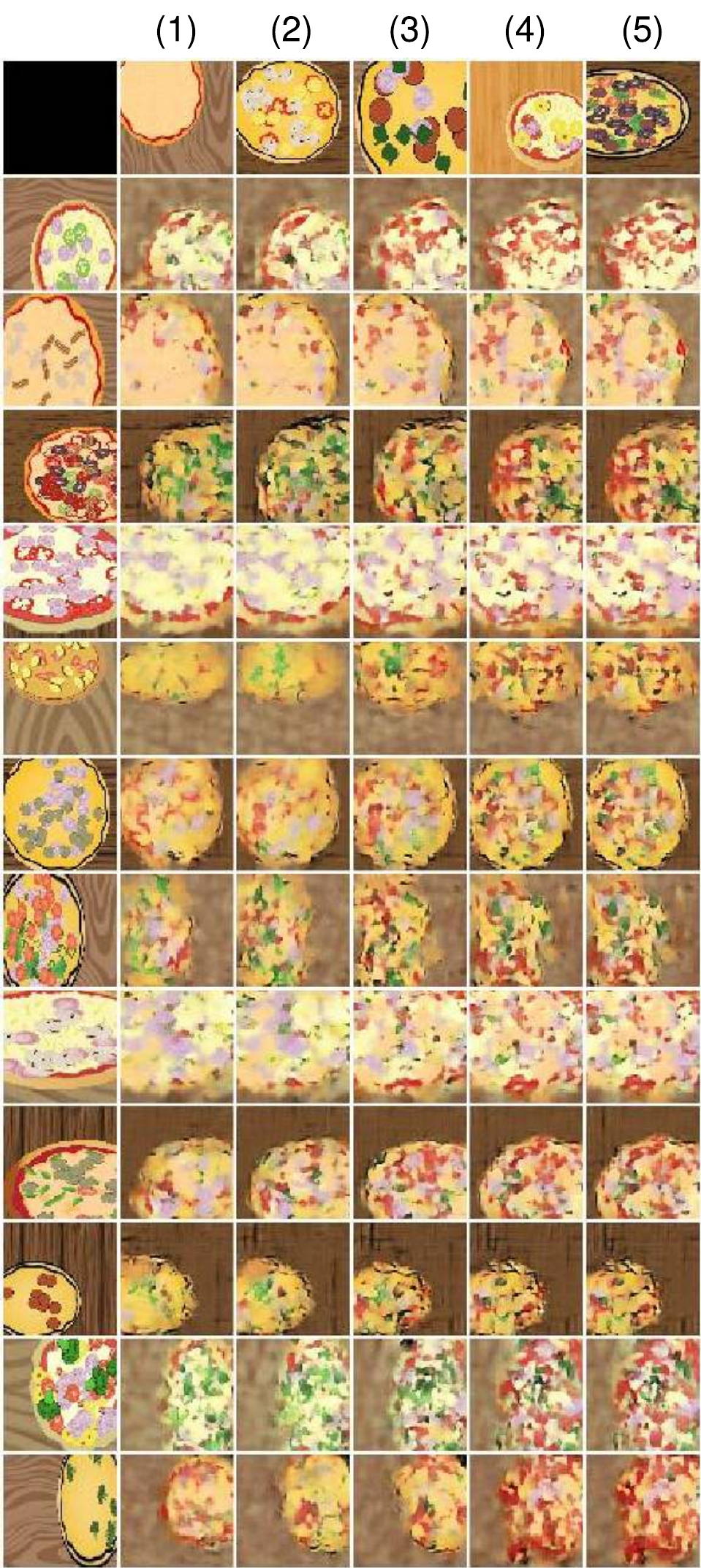}
\includegraphics[trim = 0mm 0mm 0mm 0mm, clip, scale=0.510
]{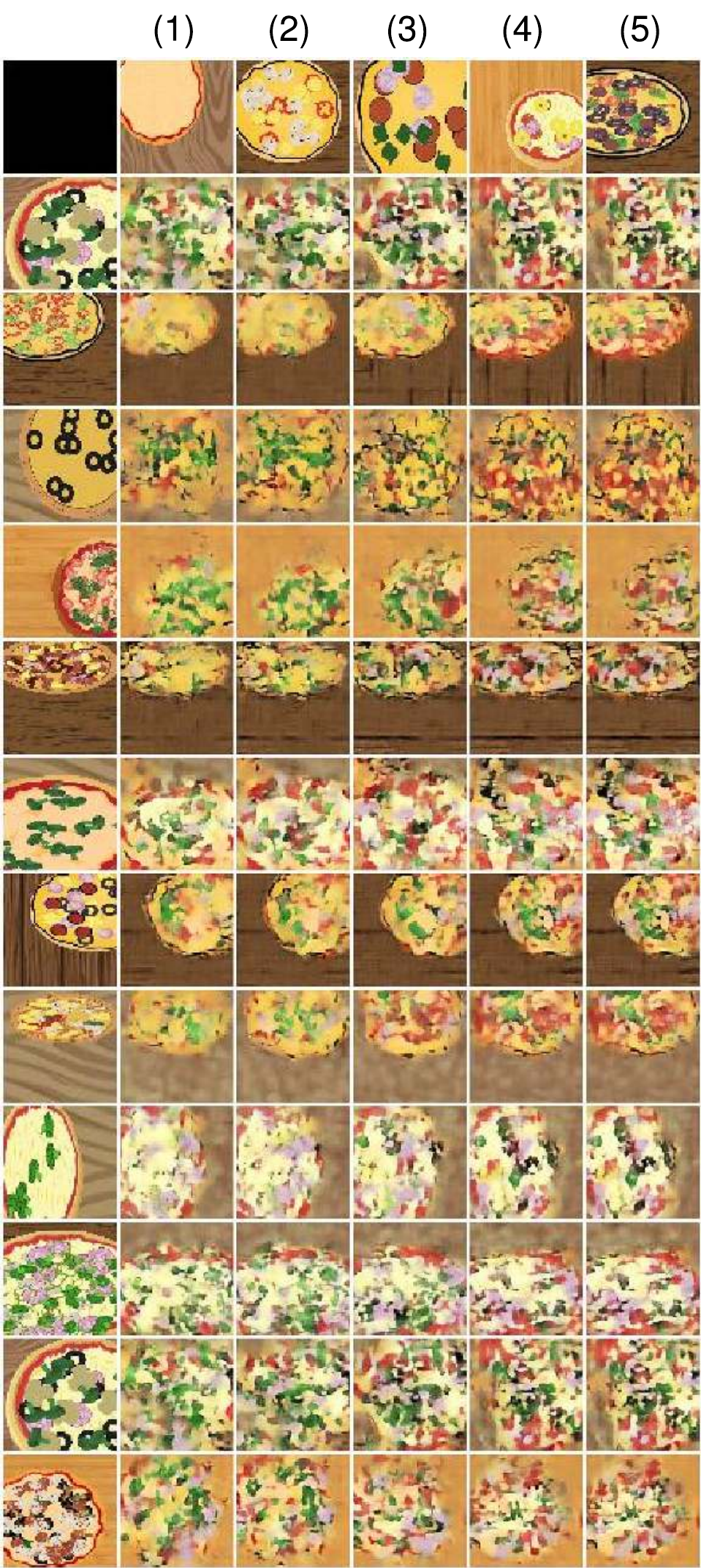}
\includegraphics[trim = 0mm 0mm 0mm 0mm, clip, scale=0.510
]{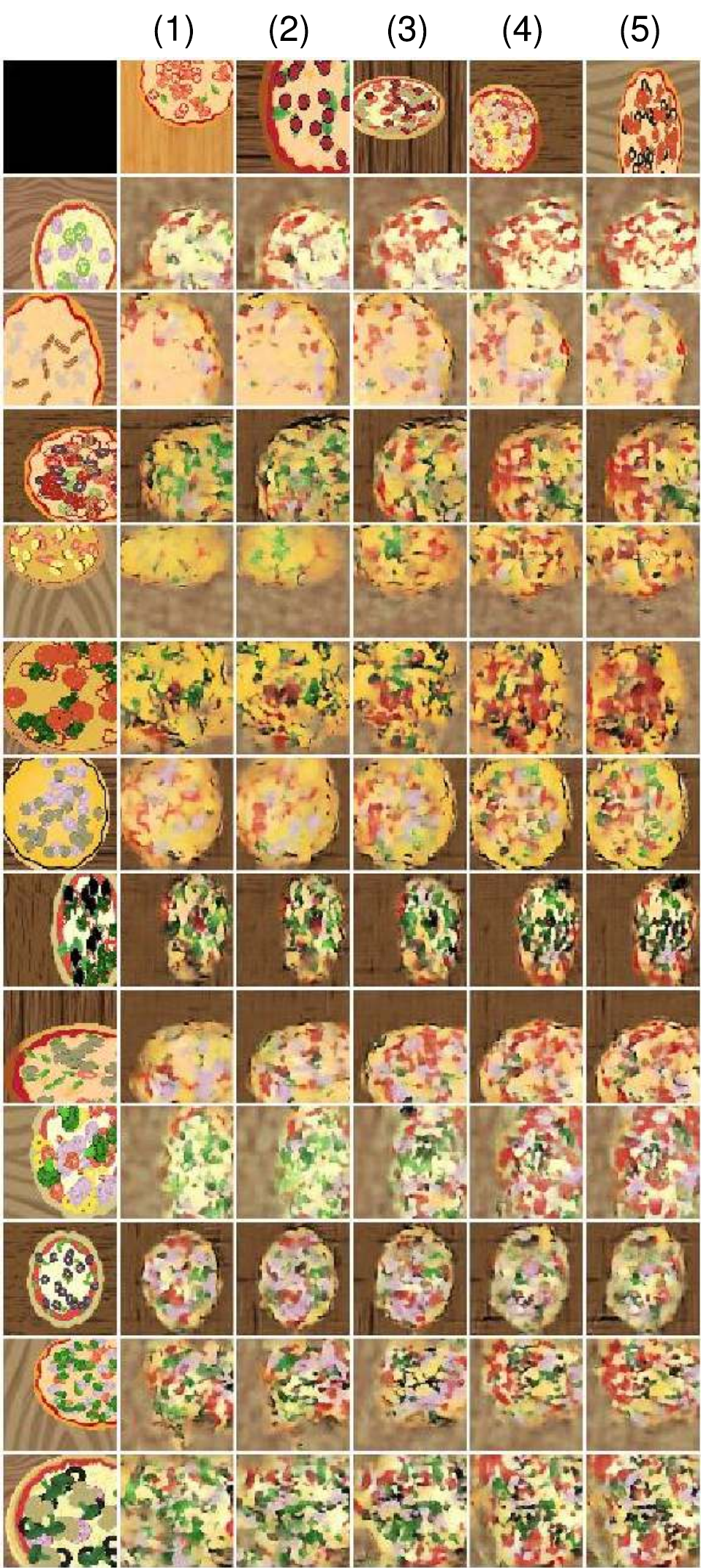}
\end{subfigure}
\end{center}
\vspace{-0.7em}
\caption{Swapping results 
of OL-VAE. 
(\textbf{Top: IMDB-WIKI}) The $K=17$ content sample images on the top are the samples with increasing content values: $c=1$ to $c=17$ (also with the age range for each group). 
(\textbf{Bottom: Pizza}) 
The $K=5$ content sample images on the top are the samples with increasing content (calorie) values: $c=1$ to $c=5$. 
For both datasets, the content vectors were inferred with $M=20$ samples per group. 
}
\label{fig:swap_imdbwiki_pizza}
\end{figure*}

\textbf{IMDB-WIKI dataset}.
To demonstrate this qualitative performance, we collect some random images ${\bf x}^\textrm{st}$ from the test set (i.e., the WIKI dataset), while selecting $17$ images ${\bf x}^\textrm{co}$ from the training set (i.e., the IMDB dataset). 
The swapping results for OLVAE and MLVAE are shown in \autoref{fig:swap_imdbwiki_pizza}(a). 
Results of our OL-VAE show that the hallucinated images largely preserve the style aspects of the reference style images (e.g., identity and pose), while conforming to the content values of the reference content images accurately (age). 

\textbf{Pizza dataset}. 
Pizza calorie (content) swapping results are shown in \autoref{fig:swap_imdbwiki_pizza}(b). The leftmost column contains reference images of style, the top row has content images, calories from low (left) to high (right). Intuitively, swapping content and style here means creating lower or higher calorie visual versions of the original pizza. 
Although the generated images look slightly blurred, the OL-VAE clearly learns important visual cues about calories such as using green-ish ingredients (vegetables and greens) for low calorie meals and red/black-ish (pepperoni/bacon and olives) for high calorie pizzas. The style aspects are relatively well preserved, notably the size and the location of the pizzas in images.

\begin{figure}
\vspace{-1.0em}
\begin{center}
\centering
\begin{subfigure}[t]{0.45\textwidth}
\centering
\includegraphics[trim = 0mm 0mm 0mm 0mm, clip, scale=0.235
]{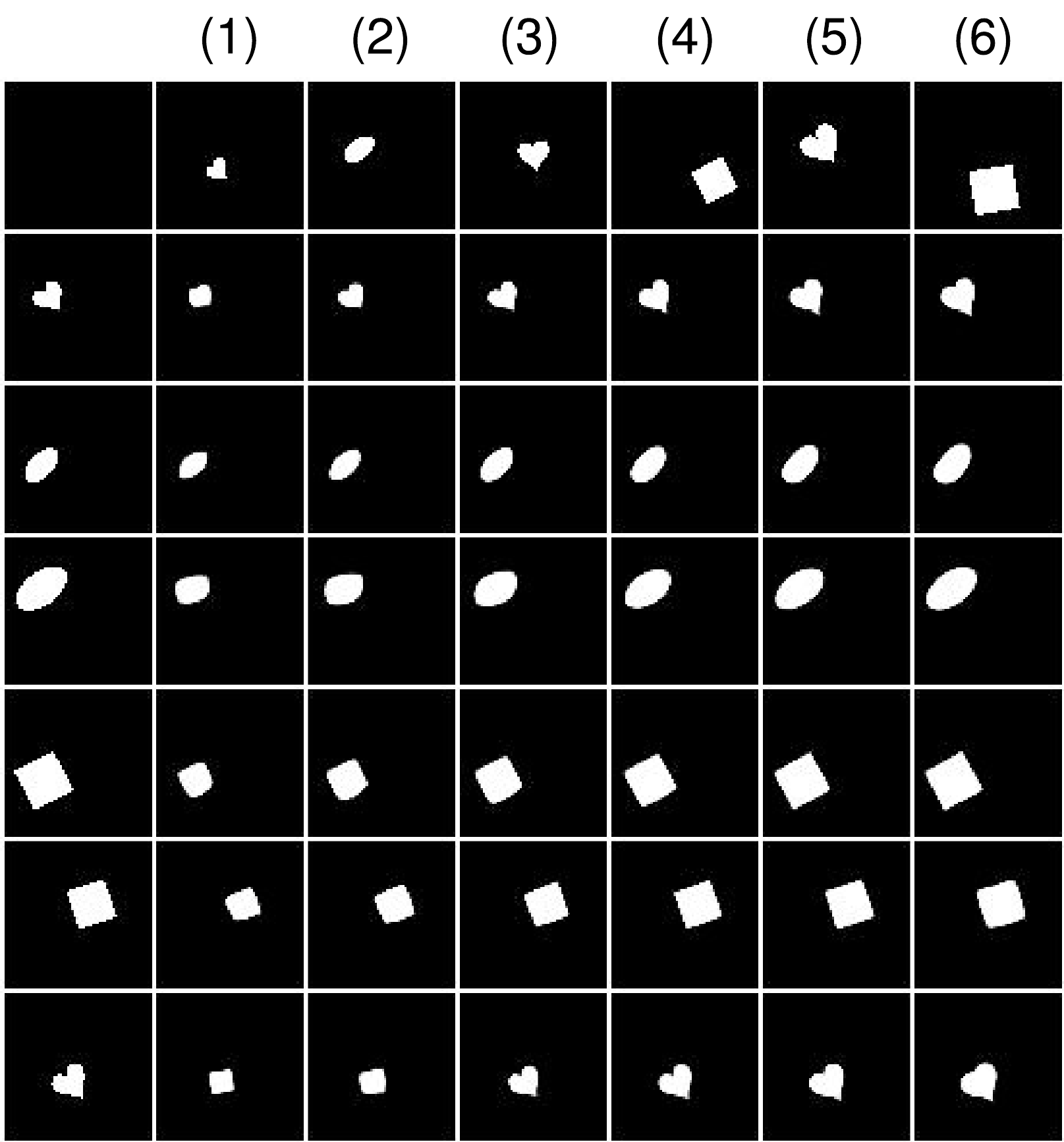} \ 
\includegraphics[trim = 0mm 0mm 0mm 0mm, clip, scale=0.235
]{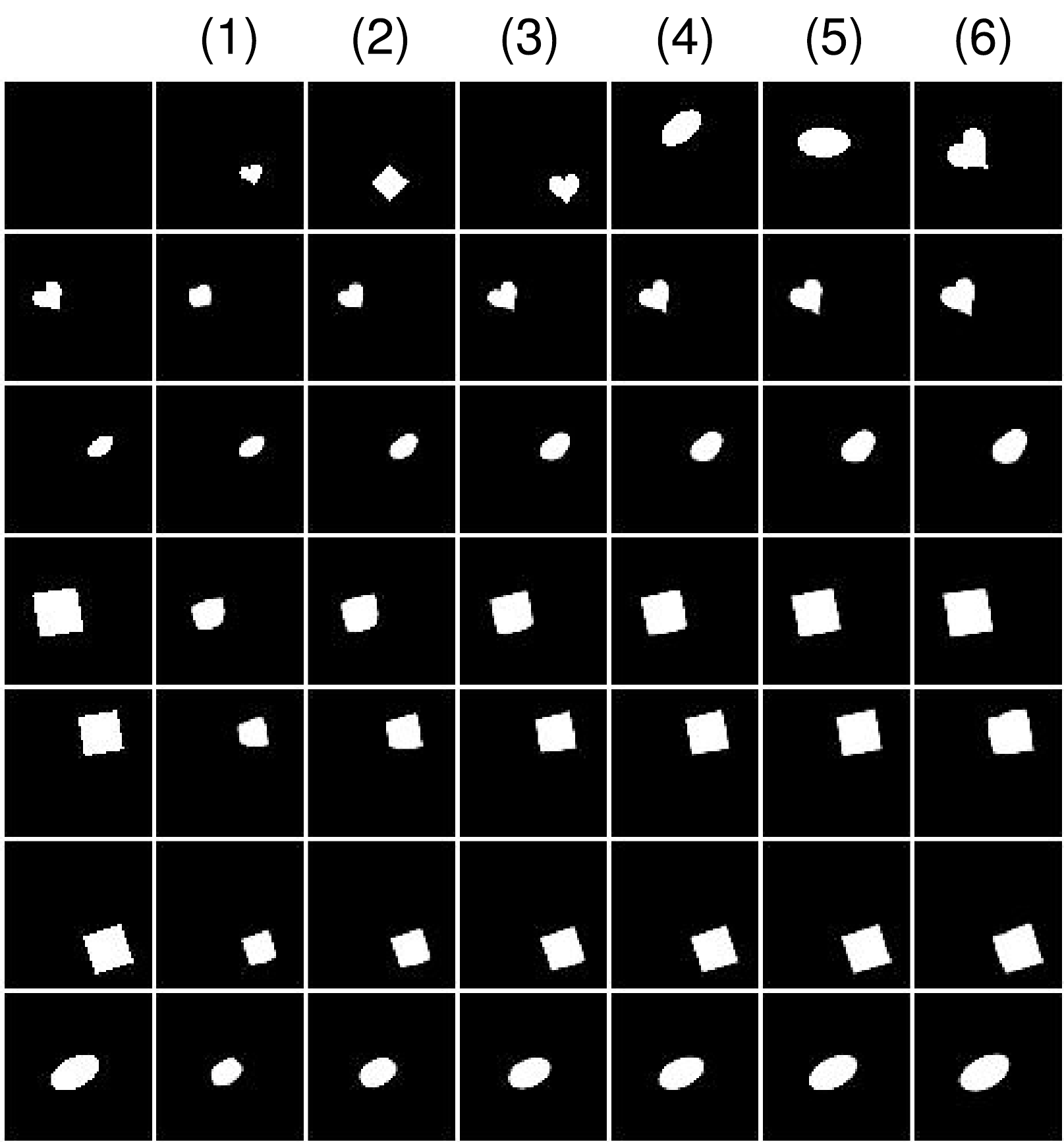} \ 
\vspace{-1.2em}
\caption{OL-VAE}
\end{subfigure} \\ \vspace{+0.5em}
\begin{subfigure}[t]{0.45\textwidth}
\centering
\includegraphics[trim = 0mm 0mm 0mm 0mm, clip, scale=0.235
]{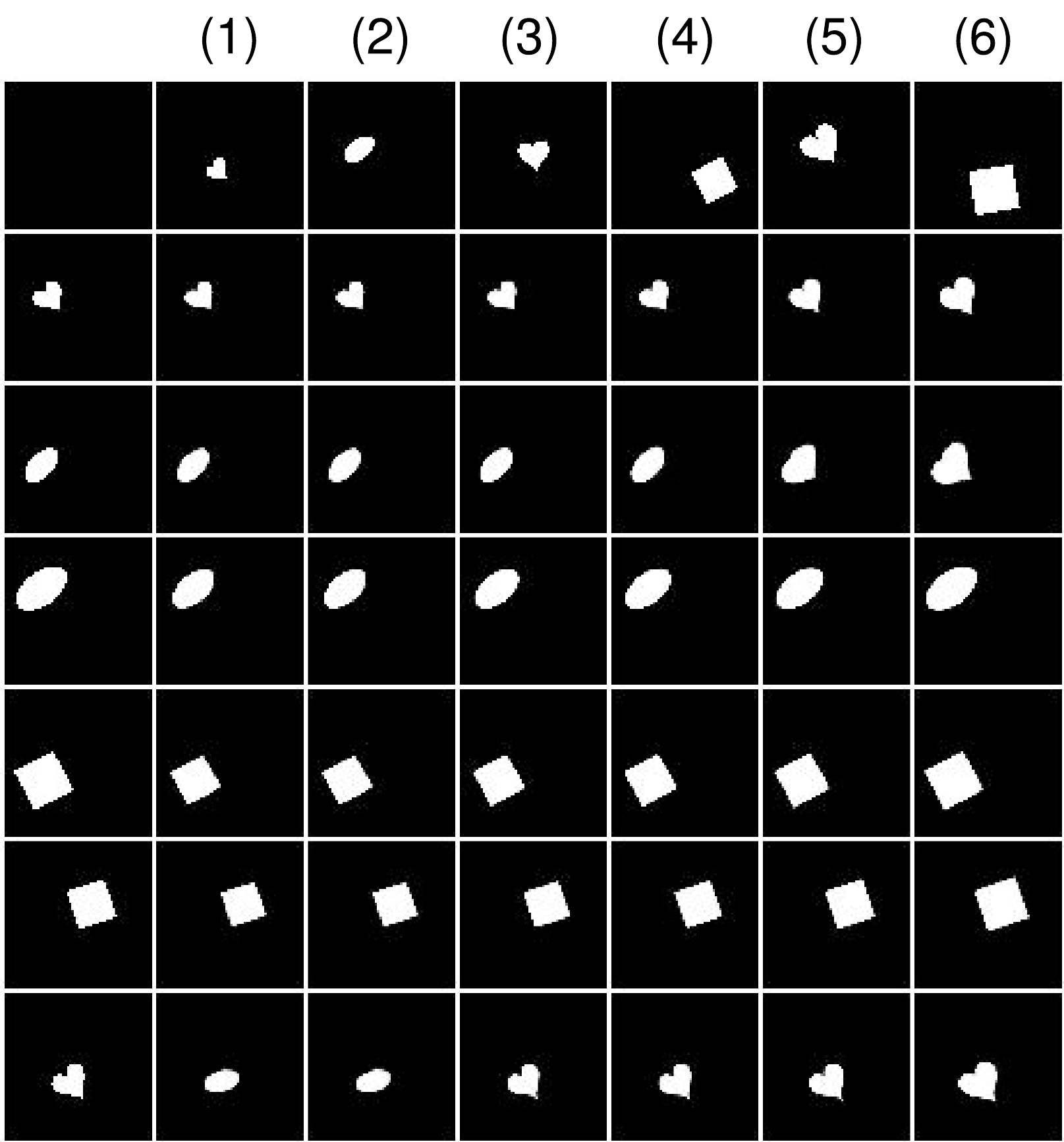} \ 
\includegraphics[trim = 0mm 0mm 0mm 0mm, clip, scale=0.235
]{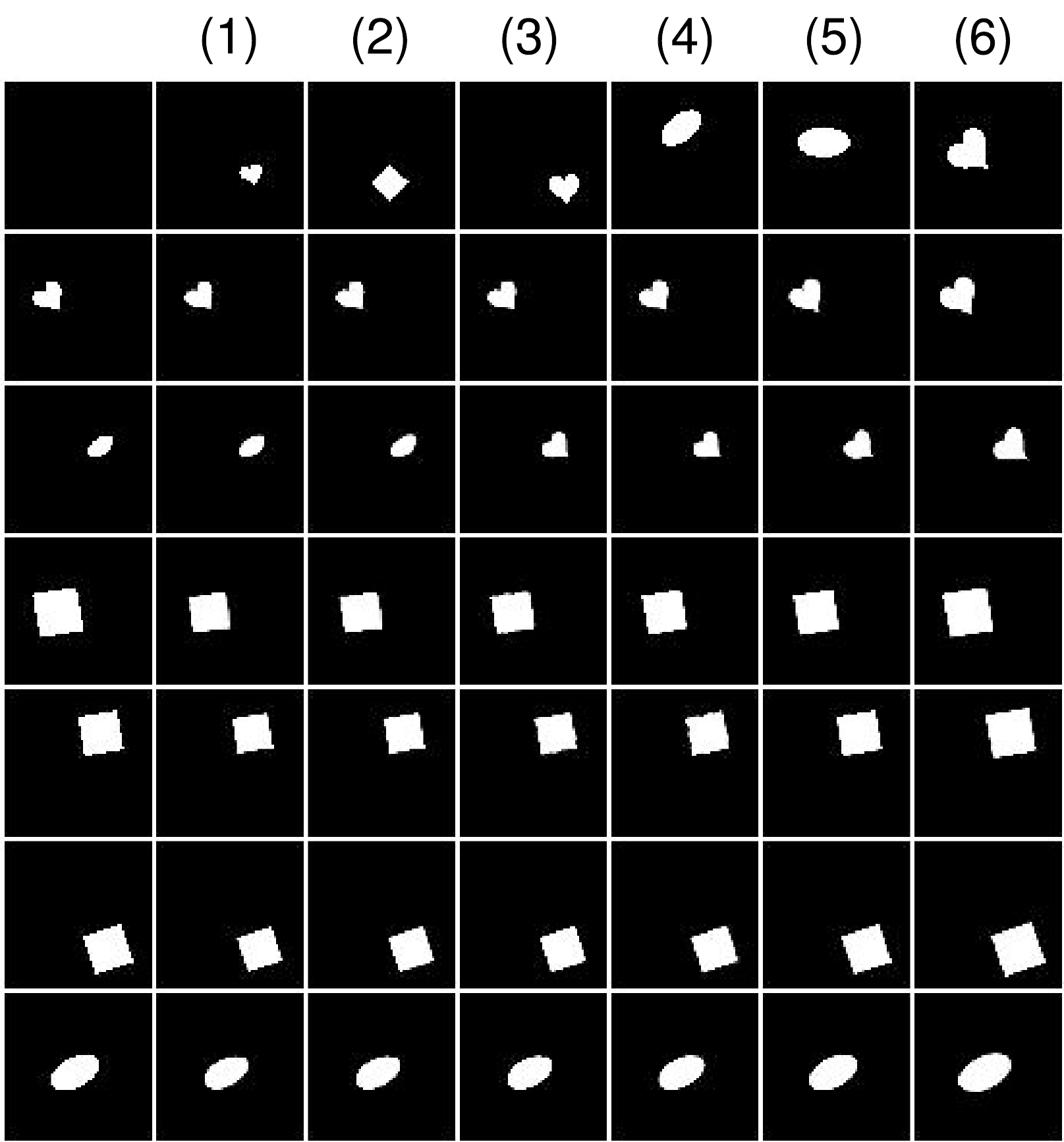} \ 
\vspace{-1.2em}
\caption{ML-VAE}
\end{subfigure}
\end{center}
\vspace{-1.0em}
\caption{Scale-Sprites swapping results of OL-VAE and ML-VAE. This is for content vector inference with $M=20$ samples per content (scale) group. The 6 content sample images on the top are the samples with increasing content values: $c=1$ to $c=6$.
}
\label{fig:swap_sprites_scale45}
\end{figure}

\begin{figure}
\vspace{-1.0em}
\begin{center}
\centering
\begin{subfigure}[t]{0.45\textwidth}
\centering
\includegraphics[trim = 0mm 0mm 0mm 0mm, clip, scale=0.235
]{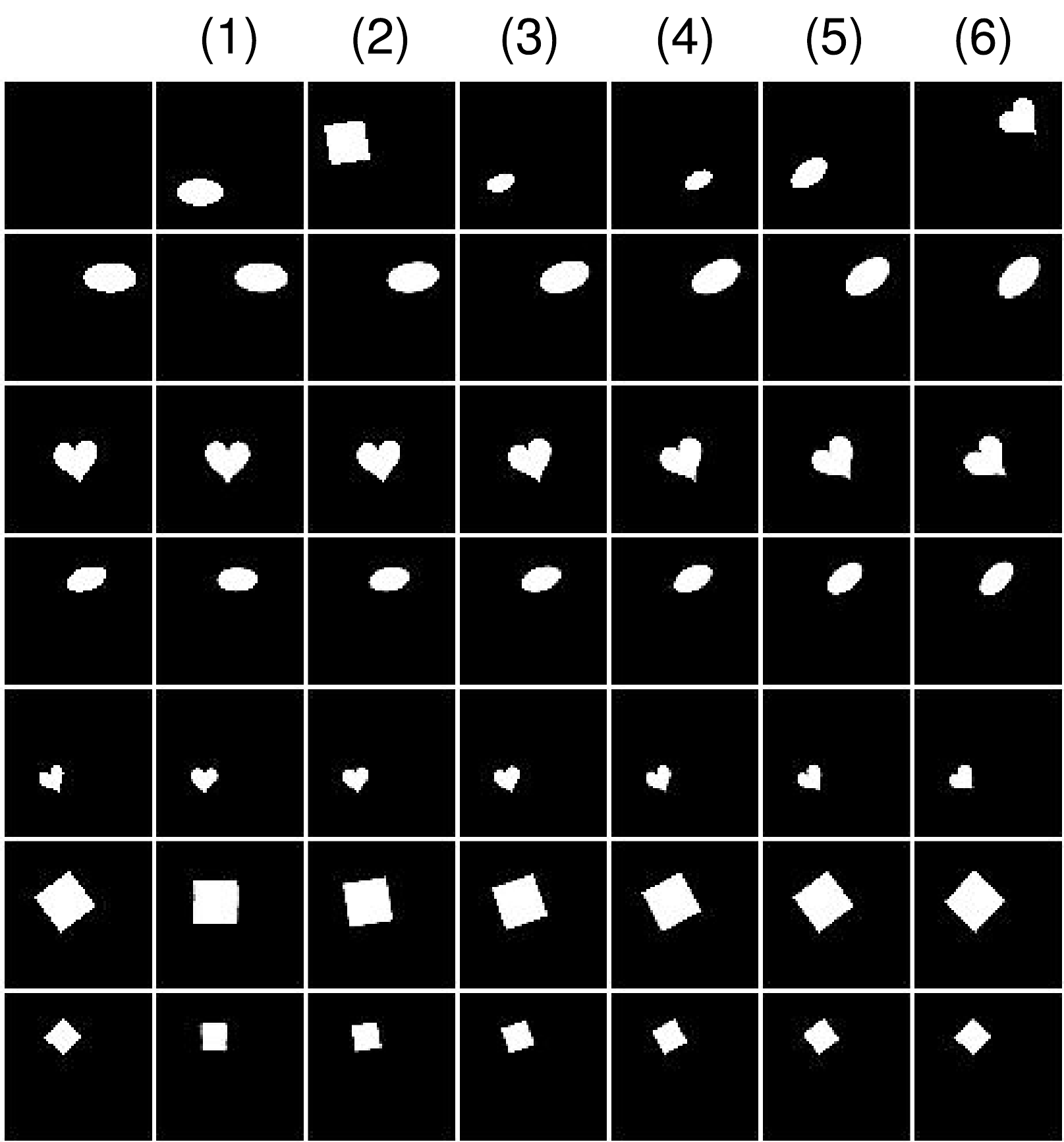} \ 
\includegraphics[trim = 0mm 0mm 0mm 0mm, clip, scale=0.235
]{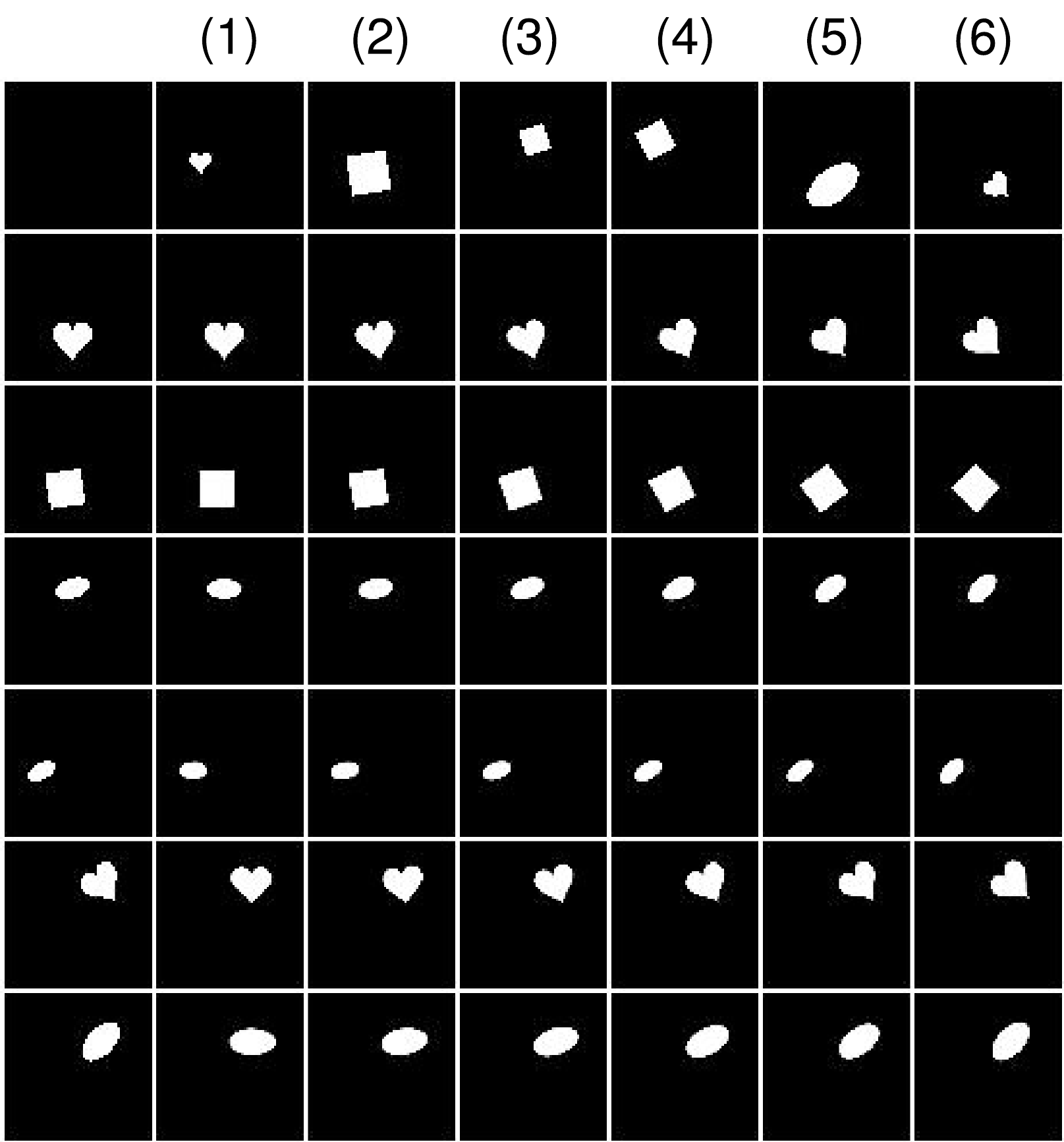} \ 
\vspace{-1.2em}
\caption{OL-VAE}
\end{subfigure} \\ \vspace{+0.5em}
\begin{subfigure}[t]{0.45\textwidth}
\centering
\includegraphics[trim = 0mm 0mm 0mm 0mm, clip, scale=0.235
]{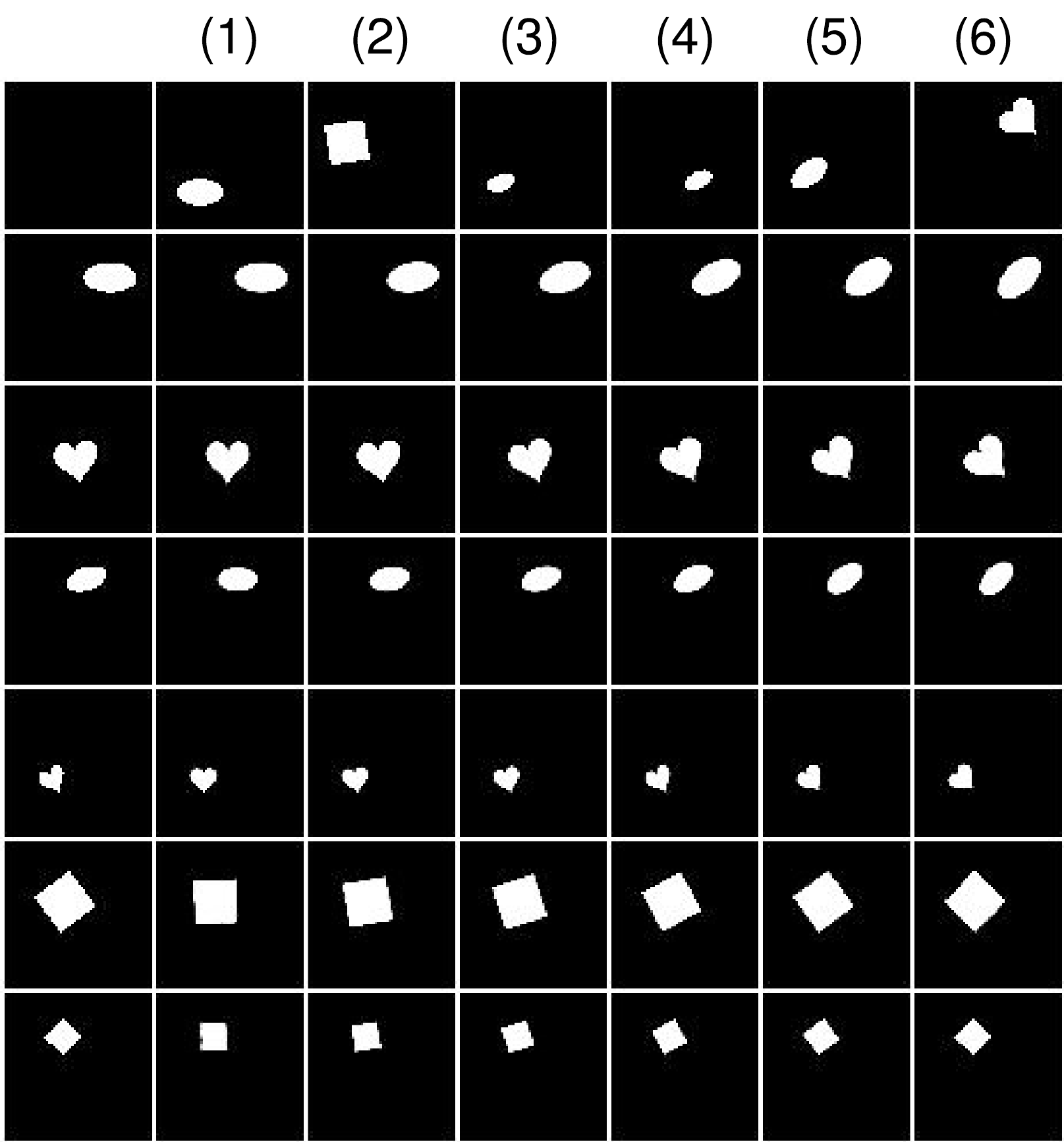} \ 
\includegraphics[trim = 0mm 0mm 0mm 0mm, clip, scale=0.235
]{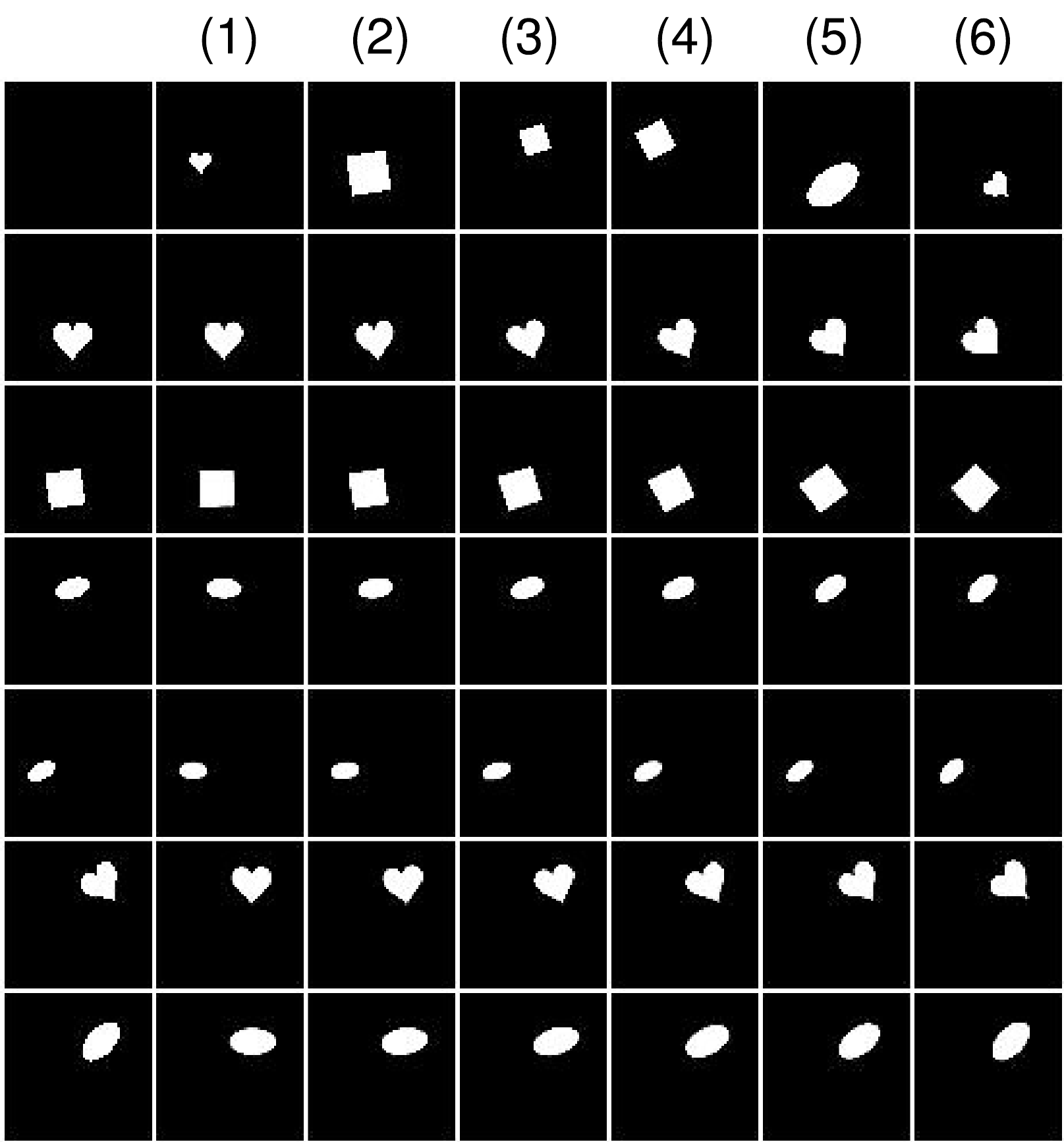} \ 
\vspace{-1.2em}
\caption{ML-VAE}
\end{subfigure}
\end{center}
\vspace{-1.0em}
\caption{Rotation-Sprites swapping results of OL-VAE and ML-VAE. This is for content vector inference with $M=20$ samples per content (rotation angle) group. The 6 content sample images on the top are the samples with increasing content values: $c=1$ to $c=6$.
}
\label{fig:swap_sprites_rot45}
\end{figure}

\textbf{Sprites datasets}. The results are shown in \autoref{fig:swap_sprites_scale45} (scale as content) and \autoref{fig:swap_sprites_rot45} (rotation as  content). On the scale sprites dataset, the OL-VAE swapping results visually have higher quality than ML-VAE. In the rotation content case, both OL-VAE and ML-VAE are equally good visually, but the classification performance in the previous section indicates OL-VAE's content-style separation is more significant.


\section{Conclusion}

Isolating content from style is a key task in deep representational learning.  In this paper we have shown that a critical step toward that goal is to provide adequate priors of the content latent space, while aligning that space with the available content label.  To that end, we focused on a specific setting where the content is ordinal (e.g., age of a person in an image). We proposed a new class of representational models, OL-VAE, characterized by a novel ordinal content space prior, the conditional Gaussian spacing model.  The prior effectively enforces the ordinal space constraints as reflected in improved separation of content from style, at the same time allowing computationally tractable learning.  To demonstrate the benefits of this model, we used 
challenging real and synthetic datasets. Quantitative content-prediction metrics and qualitative evaluations based on content-style swapping all indicate that the proposed model offers significant advantages over models that do not take into account the ordinal nature of the underlying content space.

{\small
\bibliographystyle{IEEEtran}
\bibliography{main}
}

\appendix


\subsection{Derivation of the Joint Gaussian form 
}\label{sec:joint_prior}

We provide full derivations for the joint Gaussian forms (\ref{eq:pv_joint_mean}--\ref{eq:pv_joint_ex}) of our conditional Gaussian spacing model (\ref{eq:pv_olvae_conditioned}--\ref{eq:pv_olvae_conds_4}) in the main paper. 
From the conditional model in the main paper, we have, for $i=2,\dots,K$,
\begin{equation}
P(v_i|v_{i-1}) = \mathcal{N}(v_i;\ v_{i-1}+\Delta_i, \ \sigma_i^2)
\label{eq:cond}
\end{equation}
We rewrite the above as a stochastic equation form as follows, by introducing the random variables $\epsilon_i\sim \mathcal{N}(0,\sigma_i^2)$, each of which accounts for the randomness of $v_i$ conditioned on $v_{i-1}$:
\begin{equation}
v_i = v_{i-1} + \Delta_i + \epsilon_i \ \ \ \ \textrm{for} \ \ i \geq 2
\label{eq:cond_ste}
\end{equation}
Note that $\epsilon$'s are independent with each other, and in particular $\textrm{Cov}(\epsilon_i, \epsilon_j) = 0$ if $i \neq j$. From the recursion  (\ref{eq:cond_ste}), we can write $v_i$ as (for $i \geq 2$):
\begin{equation}
v_i = \mu_1 + (\Delta_2 + \cdots + \Delta_i) + (\epsilon_1 + \cdots + \epsilon_i),
\label{eq:vi}
\end{equation}
where we use $v_1 = \mu_1 + \epsilon_1$ with $\epsilon_1 \sim \mathcal{N}(0,\sigma_1^2)$ as the initial equation.
Now, 
the mean of $v_i$ is straightforwardly derived as:
\begin{equation}
\mathbb{E}[v_i] = \mu_1 + \Delta_2 + \cdots + \Delta_i
\end{equation}
Moreover, assuming $i \leq j$ without loss of generality, 
\begin{align}
&\mathbb{\textrm{Cov}}(v_i,v_j) \ = \nonumber \\
&\ \ \ \ \ \ \mathbb{\textrm{Cov}}\big( \ 
    \mu_1 + (\Delta_2 + \cdots + \Delta_i) + (\epsilon_1 + \cdots + \epsilon_i), \nonumber \\
&\ \ \ \ \ \ \ \ \ \ \ \ \ \mu_1 + (\Delta_2 + \cdots 
    + \Delta_j) + (\epsilon_1 + \cdots 
    + \epsilon_j) \ 
    \big) \label{eq:cov_deriv_1} \\
&\ \ \ \ = \ \mathbb{\textrm{Cov}}(
    \epsilon_1 + \cdots + \epsilon_i, \ 
    \epsilon_1 + \cdots 
    + \epsilon_j) \label{eq:cov_deriv_2} \\
&\ \ \ \ = \ \mathbb{\textrm{Cov}}(\epsilon_1, \epsilon_1) + \cdots + \mathbb{\textrm{Cov}}(\epsilon_i, \epsilon_i) \label{eq:cov_deriv_3} \\
&\ \ \ \ = \ \sigma_1^2 + \cdots +  \sigma_i^2 \label{eq:cov_deriv_4} 
\end{align}
where from (\ref{eq:cov_deriv_2}) to (\ref{eq:cov_deriv_3}), we use the bilinearity of the covariance operator and mutual independence of $\epsilon$'s.

\subsection{Content Group Inference via Product of Experts}\label{sec:poe}

As described in the paper, the content latent vector can be inferred from a set of observations $\{{\bf x}^n\}_{n \in G}$ of the same content group $G$, via the product of experts rule: 
\begin{equation}
Q_c\big( {\bf v} | \{{\bf x}^n\}_{n \in G} \big) \propto \prod_{n \in G}  Q_c({\bf v} | {\bf x}^n).
\end{equation}
We derive a full formula for $Q_c\big( {\bf v} | \{{\bf x}^n\}_{n \in G} \big)$ for the Gaussian encoder network $Q_c({\bf v} | {\bf x}) = \mathcal{N}({\bf v}; {\bf m}({\bf x}), {\bf S}({\bf x}) )$ where ${\bf S}({\bf x})$ is a diagonal covariance matrix function. 

It is well known that the product of two Gaussian distributions is a Gaussian up to a constant factor, that is, 
\begin{equation}
\mathcal{N}({\bf v}; {\boldsymbol\mu}_1, {\boldsymbol\Sigma}_1) \cdot \mathcal{N}({\bf v}; {\boldsymbol\mu}_2, {\boldsymbol\Sigma}_2) \propto \mathcal{N}({\bf v}; {\boldsymbol\mu}, {\boldsymbol\Sigma}), 
\end{equation}
where ${\boldsymbol\Sigma} = \big( {\boldsymbol\Sigma}_1^{-1} + {\boldsymbol\Sigma}_2^{-1} \big)^{-1}$ and $ {\boldsymbol\mu} = {\boldsymbol\Sigma} \cdot\big( {\boldsymbol\Sigma}_1^{-1} {\boldsymbol\mu}_1 + {\boldsymbol\Sigma}_2^{-1} {\boldsymbol\mu}_2 \big)$. 
The proof for the above is from straightforward algebra that reduces to combining two quadratic exponents from the two Gaussians. 
Extension to product of multiple (more than two) Gaussians is also straightforward, which simply amounts to applying the bi-product result recursively. The posterior of the group inference can then be written as follows:
\begin{equation}
Q_c\big( {\bf v} | \{{\bf x}^n\}_{n \in G} \big) = \mathcal{N}( {\bf v}; {\bm\mu}_G, {\bm\Sigma}_G ), 
\end{equation}
where
\begin{align}
{\bm\Sigma}_G \ &= \ \Bigg( \sum_{n\in G} {\bf S}({\bf x}^n)^{-1} \Bigg)^{-1}, \\
{\bm\mu}_G \ &= \ {\bm\Sigma}_G \cdot
      \Bigg( \sum_{n\in G} {\bf S}({\bf x}^n)^{-1} {\bf m}({\bf x}^n) \Bigg).
\end{align}
    
\subsection{Variational Lower Bound (ELBO)}\label{sec:elbo}

We show that the variational lower bound (ELBO) for the data log-likelihood, can be expressed as (up to some constant):
\begin{align}
&\sum_{i=1}^K \sum_{n\in G_i} \mathbb{E}_{Q_c({\bf v}_i|G_i) Q_s( {\bf s}^n | {\bf x}^n)} \Big[ \log P({\bf x}^n|{\bf v}_i,{\bf s}^n) \Big] \ - \nonumber \\
&\ \ \textrm{KL}\Bigg( 
  \prod_{i=1}^K Q_c({\bf v}_i | G_i) \bigg\Vert P({\bf v}_1,\dots,{\bf v}_K) 
\Bigg) \ - \nonumber \\ 
&\ \ \ \sum_{n=1}^N \textrm{KL}\Big( 
  Q_s\big( {\bf s}^n | {\bf x}^n \big) \big\Vert P({\bf s}^n) 
\Big)
\label{eq:elbo}
\end{align}
To this end, we begin with the KL divergence between the true posterior distribution and our variational approximation. With the full joint distribution of our model ((\ref{eq:level_vae_full}) in the paper), 
\begin{align}
& \textrm{KL} \Bigg( \prod_{i=1}^K Q_c({\bf v}_i|G_i) \prod_{n=1}^N Q_s( {\bf s}^n | {\bf x}^n) \ \bigg\Vert \ \nonumber \\
& \ \ \ \ \ \ \ \ \ \ \ \ \ \ 
    P\big( \{{\bf v}_i\}_{i=1}^K, \{{\bf s}^n\}_{n=1}^N \ \big| \ \{(c^n,{\bf x}^n)\}_{n=1}^N \big) \Bigg) \label{vi_deriv_01} \\
&\ \ = \ \mathbb{E}_{ Q } \Bigg[
  \log \frac{ \prod_{i=1}^K Q_c({\bf v}_i|G_i) \prod_{n=1}^N Q_s( {\bf s}^n | {\bf x}^n ) }
  { P({\bf v}_1,\dots,{\bf v}_K) \prod_{i=1}^K \prod_{n\in G_i}
  P({\bf s}^n) P({\bf x}^n | {\bf v}_i, {\bf s}^n) }
\Bigg] \nonumber \\ 
& \ \ \ \ \ \ \ \ \ \ \ \ \ \ \ + \log P\big(\{(c^n,{\bf x}^n)\}_{n=1}^N \big) \ + \ \textrm{const.} & \label{vi_deriv_02} \\
&\ \ = \ \textrm{KL}\Bigg( 
    \prod_{i=1}^K Q_c({\bf v}_i | G_i) \bigg\Vert P({\bf v}_1,\dots,{\bf v}_K) 
  \Bigg) \ + \nonumber \\
& \ \ \ \ \ \ \ \log P\big(\{(c^n,{\bf x}^n)\}_{n=1}^N \big) \ + \ 
  \sum_{n=1}^N \textrm{KL}\Big( 
    Q_s\big( {\bf s}^n | {\bf x}^n \big) \big\Vert P({\bf s}^n) \Big) \nonumber \\ 
& \ \ \ \ \ \ \ - \ \mathbb{E}_{ Q } \Bigg[ \sum_{i=1}^K \sum_{n\in G_i} \log P({\bf x}^n | {\bf v}_i, {\bf s}^n) \Bigg] \ + \ \textrm{const.} \label{vi_deriv_03}
\end{align}
The last expectation term becomes identical to the first term of (\ref{eq:elbo}) due to the decomposition of the variational density $Q$. Using the fact that KL divergence (\ref{vi_deriv_01}) is non-negative, we have (\ref{eq:elbo}) as a lower bound of the data log-likelihood.

\end{document}